%% file: traveltime.tex
\pdfoutput=1

\documentclass{IEEEtran}
\usepackage{amssymb,amsmath,epsfig,graphics,enumerate,float,subfigure,verbatim,xspace, color, wrapfig, url}

\usepackage{multirow}
\usepackage{pbox}

\let\labelindent\relax
\usepackage{enumitem}
\setlist[itemize]{leftmargin=*}

\usepackage{flushend}

\newcommand{\nop}[1]{}

\newcommand{\eg}{{e.g.}}
\newcommand{\ie}{{i.e.}}

\newcommand{\etal}{{et al.}}

\newcommand{\p}{\mathbf{p}}
\newcommand{\q}{\mathbf{q}}

\newcommand{\D}{\mathcal{D}}
\newcommand{\N}{\mathcal{N}}

\newcommand{\Tr}{\mathcal{T}_{rela}}
\newcommand{\Ta}{\mathcal{T}_{abs}}

\newtheorem{problem}{Problem}
\newtheorem{assumption}{Assumption}

\newcommand{\lr}{\texttt{LR}\xspace}
\newcommand{\avg}{\texttt{AVG}\xspace}
\newcommand{\temp}{\texttt{TEMP}\xspace}
\newcommand{\temprel}{$\temp_{\texttt{rel}}$\xspace}
\newcommand{\temporacle}{$\temp_{\texttt{Oracle}}$\xspace}
\newcommand{\temparima}{$\temp_{\texttt{abs}}$\xspace}
\newcommand{\tempR}{$\texttt{TEMP+R}$\xspace}
\newcommand{\temprelR}{$\temp_{\texttt{rel}}+\texttt{R}$\xspace}

\newcommand{\temparimaR}{$\temp_{\texttt{abs}}+\texttt{R}$\xspace}
\newcommand{\seg}{\texttt{SEGMENT}\xspace}
\newcommand{\subpath}{\texttt{SUBPATH}\xspace}
\newcommand{\bing}{\texttt{BING}\xspace}
\newcommand{\bingtraffic}{\texttt{BING(traffic)}\xspace}
\newcommand{\baidu}{\texttt{BAIDU}\xspace}

\begin{document}
\title{A Simple Baseline for Travel Time Estimation using Large-Scale Trip Data}
 
\author{\IEEEauthorblockN{Hongjian Wang\IEEEauthorrefmark{1}, Zhenhui Li\IEEEauthorrefmark{1}, Yu-Hsuan Kuo\IEEEauthorrefmark{2}, Dan Kifer\IEEEauthorrefmark{2}}\\
\IEEEauthorblockA{\IEEEauthorrefmark{1}College of Information Sciences and Technology, Pennsylvania State University\\
\IEEEauthorrefmark{2}Department of Computer Science and Engineering, Pennsylvania State University\\
\IEEEauthorrefmark{1}\{hxw186,jessieli\}@ist.psu.edu, \IEEEauthorrefmark{2}\{yzk5145,dkifer\}@cse.psu.edu}
} 

\nop{
Hongjian Wang  \hspace*{0.6in}
Zhenhui Li  \hspace*{0.6in} 
Yu-Hsuan Kuo \hspace*{0.6in}
Dan Kifer  \\[1ex]
 Pennsylvania State University \\ 
 University Park, PA, USA \\
\{a, b, c\}@psu.edu
}

\maketitle

\input{0abstract}
\input{1intro-new}
\input{5relatedwork}

\input{2problem}

\input{3method-weight}

\input{3method-noise}
\input{4experiment}

\input{6conclusion}
\bibliographystyle{IEEEtran}
\bibliography{ref_jessie}
\end{document}

%% file: 0abstract.tex
\begin{abstract}
The increased availability of large-scale trajectory data around the world provides rich information for the study of urban dynamics. For example, New York City Taxi \& Limousine Commission regularly releases source/destination information about trips in the taxis they regulate~\cite{nyctaxi}. Taxi data provide information about traffic patterns, and thus enable the study of urban flow -- what will traffic between two locations look like at a certain date and time in the future?
Existing big data methods try to outdo each other in terms of complexity and algorithmic sophistication. In the spirit of ``big data beats algorithms'', we present a very simple baseline which outperforms state-of-the-art approaches, including Bing Maps~\cite{bingmap} and Baidu Maps~\cite{baidumap} (whose APIs permit large scale experimentation).
Such a travel time estimation baseline has several important uses, such as navigation (fast travel time estimates can serve as approximate heuristics for $A^*$ search variants for path finding) and trip planning (which uses operating hours for popular destinations along with travel time estimates to create an itinerary).



\end{abstract}

%% file: 1intro-new.tex
\section{Introduction}
\label{sec:intro}

The positioning technology is widely adopted into our daily life. The on-board GPS devices track the operation of vehicles and provide navigation service, meanwhile a significant amount of trajectory data are collected. For example, the New York City Taxi \& Limousine Commission has made all the taxi trips in 2013 public under the Freedom of Information Law (FOIL) \cite{nyctaxi}. The dataset contain more than 173 million taxi trips -- almost half million trips per day. These trajectory data can help us in a lot of ways. In the macro scale, we can study the urban flow, while in the micro level, we can predict the travel time for individual users.

\begin{figure}[b!]
\centering
\includegraphics[width=0.35\textwidth]{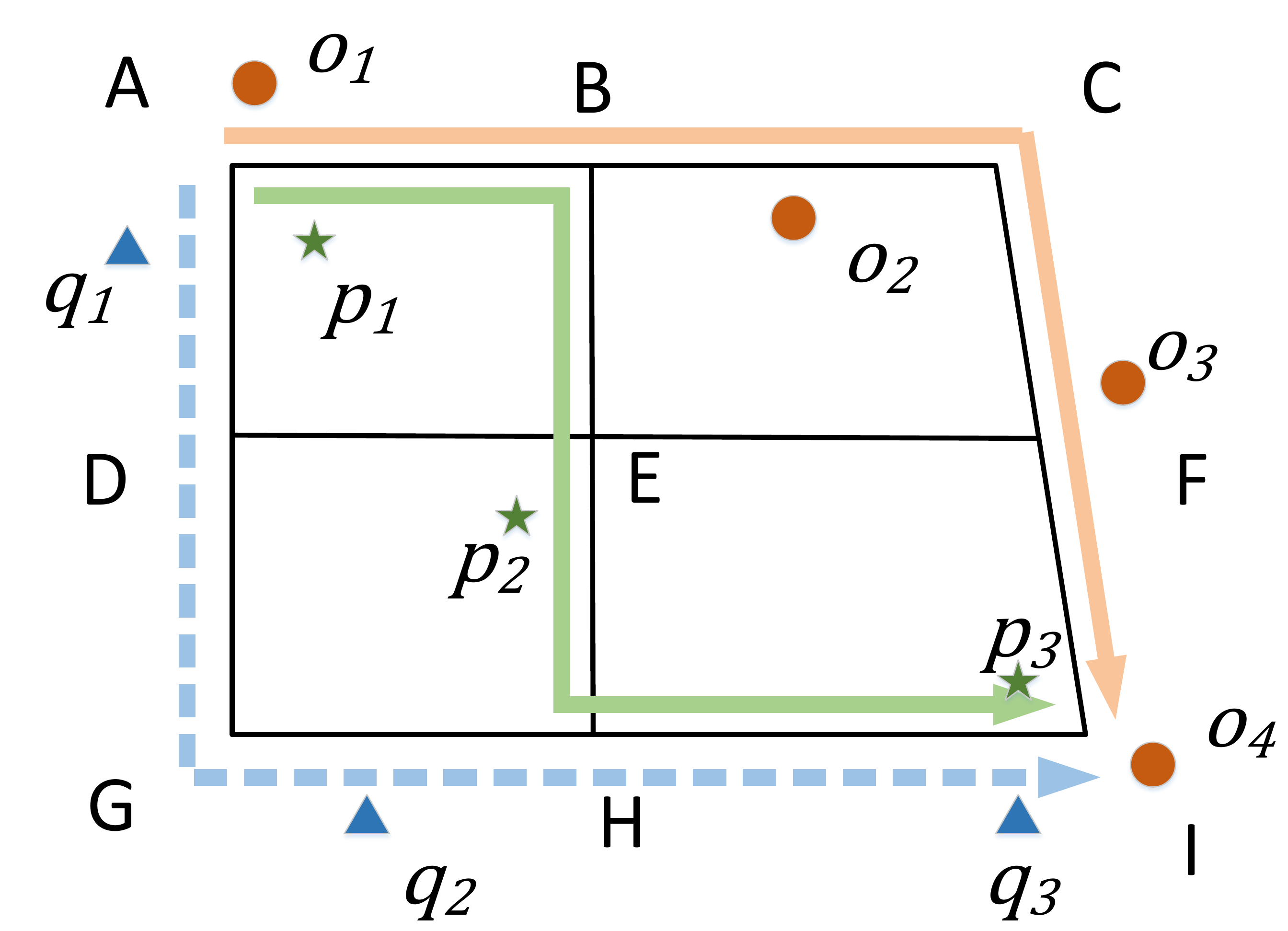}
\caption{An example of travel time estimation. We want to estimate the travel time from point $A$ to point $I$. There are three historical trips available, which are trip $\mathbf{o}$, $\mathbf{p}$, and $\mathbf{q}$. All three trips take different route traveling to $I$ from $A$.}
\label{fig:examp}
\end{figure}

In this paper, we focus on a fundamental problem, which is the travel time estimation -- predicting the travel time between an origin and a destination. Existing big data methods try to outdo each other in terms of complexity and algorithmic sophistication. In the spirit of ``big data beats algorithms'', we present a very simple baseline which outperforms state-of-the-art approaches, including Bing Maps~\cite{bingmap} and Baidu Maps~\cite{baidumap}. Figure~\ref{fig:examp} gives an example of a road network and three given trajectories $\mathbf{o}, \mathbf{p}$, and $\mathbf{q}$. Suppose we want to estimate the travel time from point $A$ to $I$ using these three historical trajectories. Most existing methods of travel time estimation use the \emph{route-based framework} to estimate the travel time \cite{Dean99, KDXZ06, Gonz+07, Yuan+10}. At the first step, they usually find a specific route according to some criteria, examples are shortest route and fastest route. In the example, suppose we identify the route $A\rightarrow B\rightarrow E\rightarrow H\rightarrow I$ as the optimal route. The second step is to estimate the travel time for each segment of the chosen route. For example, the travel time of $A\rightarrow B$ is estimated from $\mathbf{o}$ and $\mathbf{p}$, since both trajectories contain this segment. Similarly, $B\rightarrow E$ and $E\rightarrow H$ are estimated from $\mathbf{p}$, and $H\rightarrow I$ can be estimated from $\mathbf{p}$ and $\mathbf{q}$.

The route-based travel time estimation faces several issues. 1) GPS data have limited precision. Usually the GPS samples are not perfectly aligned with the road segments. For a historic trip, the exact path is very hard to recover, and various intuitions are employed to guess the real route. For example, in Figure~\ref{fig:examp} trip $\mathbf{p}$ could also take the route $ADEHI$, if $p_2$ is mapped onto the road segment $DE$. 2) Trajectory data are sparse. Even with millions of trip observations, there are a lot of road segments having no GPS samples available at all, due to the non-uniform spatial distribution of GPS samples. In the Manhattan, we observe $90\%$ of taxi pick-up and drop-off concentrate within middle and south part. The inference of travel time is not accurate due to limited number of trajectories covering the road segment. In addition to the spatial sparseness, the trajectory data are temporally sparse too. Even if a road segment is covered by a few trajectories, these trajectories may not be sufficient to estimate the dynamic travel time that varies all the time.

To address the sparsity issue, various methods are proposed, which however suffer from the complexity issue. A probabilistic  framework to estimate the probability distribution of travel time is proposed in \cite{HHAB12}, where a dynamic Bayesian network is employed to learn the dependence of travel time on road congestion. Learning the  Bayesian network is NP-hard, and time complexity is exponential in the number of hidden variables. A more recent work \cite{WZX14} models the travel time over each road segment during different periods for different users as a tensor. And it addresses the missing observations issue with tensor decomposition, which is known to be NP-hard \cite{Vav09}. Is all the complexity necessary or can a simple method, more reliant on data, perform just as well or even better?

In this paper, we propose a simple baseline method to predict the travel time. Since we have a huge amount of historical trips, we can avoid finding a specific route by directly using all trips with similar origin and destination to estimate the query trip. We call our method as \emph{neighbor-based method}. The biggest benefit of our approach is its simplicity. By analogy, consider the area of multidimensional database indexing and similarity search. While many sophisticated indexing algorithms have been proposed, many are outperformed by a simple sequential search of the data \cite{WSHB98}. In the case of navigation and trip time estimation, there are no suitable baselines that set the bar for accuracy and help determine whether additional algorithmic complexity is justified. Thus, our method fills this void while, at the same time, being more accurate than the current state of the art.

To explain the intuition behind our neighbor-based approach and the reasoning of why it should outperform sophisticated competitors, consider the following scenario. Suppose we are interested in how long it takes to travel from $A$ to $I$ at 11:43 p.m. on a Saturday night, given the information in Figure~\ref{fig:examp}. If we had hundreds of trips from $A$ to $I$ starting exactly at 11:43 p.m. on Saturday, the estimation problem would be easy: just use the average of all those trips. Clearly, if such data existed, then this approach would be more accurate than the route-based method, which reconstructs a path from $A$ to $I$, estimates the time along each segment, and finally adds them up. This latter approach requires multiple accurate time estimates and any errors will naturally accumulate.

On the other hand, the simple average of neighboring trips approach suffers from sparsity issue: although taxi data contain hundreds of millions of taxi trips, very few of them share the similar combination of origin, destination, and time. To address this issue, we propose to model the dynamic traffic conditions as a temporal speed reference, which enables us to average all trips from $A$ to $I$ during different times. More specifically, for any two trips $\mathbf{p}$ and $\mathbf{q}$ from different time but having similar origin and destination, we can estimate the travel time of $\mathbf{p}$ by adjusting the travel time of $\mathbf{q}$ with the temporal speed reference. For example, if historical trip $\mathbf{p}$ is at 6 p.m. and query trip $\mathbf{q}$ is at 1 p.m.; and the data tells us travel time at 6 p.m. is usually twice slower than the travel time at 1 p.m., we could estimate travel time of $\mathbf{p}$ using that of $\mathbf{q}$ with a scaling factor of $2$.

We conduct experiments on two large datasets from different countries (NYC in US and Shanghai in China). We evaluate our method using more than $150$ million Manhattan trips, and our method is able to outperform the Bing Maps~\cite{bingmap} by $33\%$ (even when Bing uses traffic conditions corresponding to the time and the day of week of target trips)\footnote{Google Maps API rate caps prevent proper large-scale comparisons.}. Since NYC taxi data only contain the information on endpoints of the trips (i.e., pickup and drop off locations), to compare our method with route-based method \cite{WZX14}, we use Shanghai taxi data, where more than $5$ million trips with complete information of trajectories are available. On this dataset, our method also significantly outperforms the state-of-the-art route-based method \cite{WZX14} by $19\%$ and Baidu Maps\footnote{Baidu Maps is more suitable to query trips in China due to restrictions on geographic data in China~\cite{chinaGeoRestrict}.}~\cite{baidumap} by $17\%$. It is noteworthy that our method runs $40$ times faster than state-of-the-art method \cite{WZX14}.


One may argue that our simple method does not provide the specific route information, which limits its application. While this is true, it does not keep our method from being a good baseline for travel time estimation problem. Besides, there are many real scenarios where the route information is not important. One example is the trip planning \cite{LCH+05}. For example, a person will take a taxi to catch a flight, which leaves at 5pm. Since he is not driving, the specific route is less of a concern and he only needs to know the trip time. The second example is to estimate the city commuting efficiency \cite{Nie06} \cite{MK10} in a future time. In the field of urban transportation research, the commuting efficiency measure is a function of observed commuting time and expected minimum commuting time of given origin and destination areas. To estimate the commuting efficiency of a city, the ability to predict the pairwise travel time for any two areas is crucial. In this scenario, the specific path between two areas is not important, either.

In summary, the contributions of this paper are as follows:
\begin{itemize}
\item We propose to estimate the travel time using neighboring trips from the large-scale historical data.  To the best of our knowledge, this is the first work to estimate travel time without computing the routes.
\item We improve our neighbor-based approach by addressing the dynamics of traffic conditions and by filtering the bad data records.
\item Our experiments are conducted on large-scale real data. We show that our method can outperform state-of-the-art methods as well as online map services (Bing Maps and Baidu Maps).
\end{itemize}

The rest of the paper is organized as follows. Section~\ref{sec:relatedwork} reviews related work. Section~\ref{sec:problem} defines the problem and provides an overview of the proposed approach.  Sections~\ref{sec:neighbor} and~\ref{sec:noise} discuss our method to weight the neighboring trips and our outlier filtering method, respectively. We present our experimental results in Section~\ref{sec:exp} and conclude the paper in Section~\ref{sec:conclusion}.  

%% file: 5relatedwork.tex
\section{Related Work}
\label{sec:relatedwork}

To the best of our knowledge, most of the studies in literature focus on the problem of estimating travel time for \emph{a route} (\ie, a sequence of locations).  There are two types of approaches for this problem: \emph{segment-based method} and \emph{path-based method}. 

\smallskip
\emph{Segment-based method.} A straightforward travel time estimation approach is to estimate the travel time on \emph{individual road segments} first and then take the sum over all the road segments of the query route as the travel time estimation. There are two types of data that are used to estimate the travel time on road segments: \emph{loop detector data}  and  \emph{floating-car data}  (or probe data)~\cite{TrKe13}. Loop detector can sense whether a vehicle is passing above the sensor. Various methods~\cite{Pett+98, JCCV01, RiVa04, WaPa05} have been proposed to infer vehicle speed from the loop sensor readings and then infer travel time on individual road segments. Loop detector data provide continuous speed on selected road segments  with sensors embedded. 

Floating cars collect timestamped GPS coordinates via GPS receivers on the cars. The speed of individual road segments at a time $t$ can be inferred if a floating car is passing through the road segment at that time point~\cite{Work+08,Cint+13}. Due to the low GPS sampling rate, a vehicle typically goes through multiple road segments between two consecutive GPS samplings. A few methods have been proposed to overcome the low sampling rate issue~\cite{DRV08, HHAB09, HHAB12, WWMH13, JeKo13}. Another issue with floating-car data is data sparsity -- not all road segments are covered by vehicles all the time. Wang~\etal~\cite{WZX14} proposes to use matrix factorization to estimate the missing values on individual road segments. In our problem setting, we assume the input is a special type of the floating-car data, where we only know the origin and destination points of the trips. Since the trips could vary from a few minutes to an hour, it will be much more challenging to give an accurate speed estimation on individual road segments using such limited information.

\smallskip
\emph{Path-based method.} Segment-based method does not consider the transition time between road segments such as waiting for traffic lights and making left/right turns. Recent research works start considering the time spent on intersections~\cite{Herr10, HoBa11, HHAB12, LAS15}.
However, these methods do not directly use sub-paths to estimate travel time. Rahmani et al.~\cite{RJK13} first proposes to concatenate sub-paths  to give a more accurate estimation of the query route.  Wang~\etal~\cite{WZX14} further improves the path-based method by first mining frequent patterns~\cite{SSBZ14} and then concatenating frequent sub-paths by minimizing an objective function which balances the length and support of the sub-paths. Our proposed solution can be considered as an extreme case of the path-based method, where we use the travel time of the \emph{full paths} in historical data to estimate the travel time between an origin and a destination. Full paths are better at capturing the time spent on all intersections along the routes, assuming we have enough full paths (i.e., a reasonable support).

\medskip
\emph{Origin-destination travel time estimation.} In our problem setting, instead of giving a route as the input query, \emph{we only take the origin and destination as the input query}. To the best of our knowledge, we are the \emph{first} to directly work on such origin-destination (OD) travel time queries. One could argue that, we could turn the problem into the trajectory travel time estimation by first finding a route (\eg, shortest route or fastest route)~\cite{Dean99, KDXZ06, Gonz+07, Yuan+10} and then estimating the travel time for that route. This alternative solution is similar to those provided by online maps services such as Google Maps~\cite{googlemap}, Bing Maps~\cite{bingmap} or Baidu Maps~\cite{baidumap}, where users input a starting point and an ending point, and the service generates a few route options and their corresponding times. Such solution eventually leads to travel time estimation of a query route. However, the historical trajectory data may only have the starting and ending points of the trips, such as the public NYC taxi data~\cite{nyctaxi} used in our experiment. Different from the data used in literature which has the complete trajectories of floating cars, such data have limited information and are not suitable for the segment-based method or path-based method discussed above.  In addition, route-based approach introduces more expensive computations because we need to find the route first and then compute travel time on segments or sub-paths. More importantly, such expensive computation does not necessarily lead to better performance compared with our method using limited information, as we will demonstrate in the experiment section.

%% file: 2problem.tex
\section{Problem Definition}
\label{sec:problem}

A \textbf{trip} $\p_i$ is defined as a 5-tuple $(o_i,d_i,s_i, l_i, t_i)$, which consists of the origin location $o_i$, the destination location $d_i$ and the starting time $s_i$. Both origin and destination locations are GPS coordinates. We use $l_i$ and $t_i$ to denote the distance and travel time for this trip, respectively. Note that, here we assume the intermediate locations of the trips are not available. In the real applications, it is quite possible that we can only obtain such limited information about trips due to privacy concerns and tracking costs. To the best of our knowledge, the largest-scale \emph{public} trip data we can obtain are recently released NYC taxi data~\cite{nyctaxi}. And due to the privacy concerns of passengers and drivers, no information of intermediate GPS points is released (or even recorded).

\begin{problem}[OD Travel Time Estimation]
  Suppose we have a database of trips, $\D=\{ \p_i\}_{i=1}^N$. Given a query $\q=(o_q,d_q,s_q)$, our goal is to estimate the travel time $t_q$ with given origin $o_q$, destination$d_q$, and departure time $s_q$,  using the historical trips in $\D$.
\end{problem}


\subsection{Approach Overview}
\label{sec:method:problem:over}
An intuitive solution is that we should find similar trips as the query trip $\q$ and use the travel time of those similar trips to estimate the travel time for $\q$. The problem can be decomposed to two sub-problems: (1) how to define similar trips; and (2) how to aggregate the travel time of similar trips. Here, we name similar trips as \textbf{neighboring trips} (or simply \textbf{neighbors}) of query trip $\q$. Note that aggregating is not trivial because of varying starting times and traffic conditions. 

We call trip $\p_i$ a neighbor of trip $\q$ if the origin (and destination) of $\p_i$ are spatially close to the origin (and destination) of $\q$. Thus, the set of neighbors of $\q$ is defined as: 
\begin{equation}
  \N(\q) = \{ \p_i \in \D | dist(o_i, o_q)\leq \tau \text{ and } dist(d_i,d_q)\leq \tau  \},
\label{eq:neighbor}
\end{equation}
where $dist()$ is the Euclidean distance of two given points.

With the definition of neighbors, a baseline approach is to take the average travel time of these trips as the estimation:
\begin{equation}
\label{eq:prd0}
\widehat{t}_q = \frac{1}{|\N(\q)|}\sum_{\p_i\in \N(\q)} t_i.
\end{equation}

 Here we also point out an alternative definition of neighbors. Specifically, instead of using a \emph{hard} distance threshold $\tau$, we could consider a wider range of trips with weights based on the origin distance (i.e., $dist(o_i, o_q)$) and destination distance (i.e., $dist(d_i, d_q)$). For example, let $w_i \propto \displaystyle\frac{1}{dist(o_i, o_q)}$ denote the weight coefficient for neighboring trip $\p_i$, we have:
\begin{equation}
  \widehat{t}_q = \frac{\sum_{\p_i\in \N(\q)} w_i \cdot t_i}{\sum_{\p_i\in \N(\q)} w_i }.
\label{eq:model}
\end{equation}
However, such definition will make the computation more expensive since we typically need to consider more neighbors. In addition, we empirically find out that this alternative definition does not necessarily lead to a better performance (refer to Section~\ref{subsec:nyc-spatial-weight}). So we adopt our original definition of neighbors using a hard distance threshold.

With the model in Equation~\ref{eq:prd0}, there are two major issues with the baseline approach. First, we need to model the dynamic traffic condition across different time. For each neighboring trip $\p_i$ of $\q$ we define the \emph{scaling factor} $s_i$ calculated from the speed reference, so that $s_i t_i \approx t_q$. And our estimation model becomes
\begin{equation}
  \widehat{t}_q = \frac{1}{|\N(\q)|}\sum_{\p_i\in \N(\q)} s_i t_i.
  \label{eq:prd1}
\end{equation}

Second, the data could contain a large portion of noises. It is critical to filter those outliers. Otherwise, the estimation for the query trip will be severely impacted by the noises. 
Section~\ref{sec:neighbor} will discuss how to calculate scaling factors to the neighbors and Section~\ref{sec:noise} will present our outlier filtering technique.

%% file: 3method-weight.tex
\section{Capturing the Temporal Dynamics of Traffic Conditions}
\label{sec:neighbor}

\begin{figure}[b]
\centering
\includegraphics[width=0.45\textwidth]{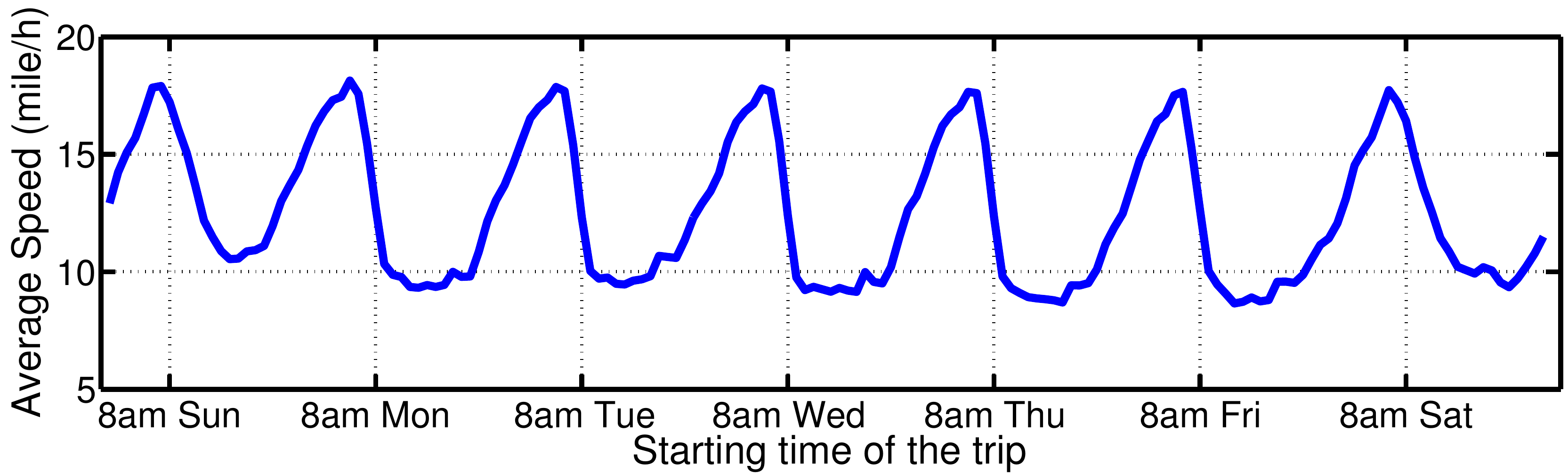}
\caption{Average travel speed w.r.t. trip starting time.}
\label{fig:method:speed}
\end{figure}

As we discussed earlier, it is not appropriate to simply take the average of all the neighbors of $\q$ because of traffic conditions vary at different times. Figure~\ref{fig:method:speed} shows the average speed of all NYC taxi trips at different times in a week. Apparently, the average speed is much faster at the midnight compared with the speed during the peak hours. Thus, if we wish to estimate the travel time of a trip $\q$ at 2 a.m. using a neighboring trip $\p$ at 5 p.m., we should proportionally decrease the travel time of $\p$, because a 2 a.m. trip is often much faster than a 5 p.m. trip. 

\begin{figure}[t]
\centering
\subfigure[$\frac{t_q}{t_i} \approx \frac{v_i}{v_q}$\label{fig:assump1}]{\includegraphics[width=0.23\textwidth]{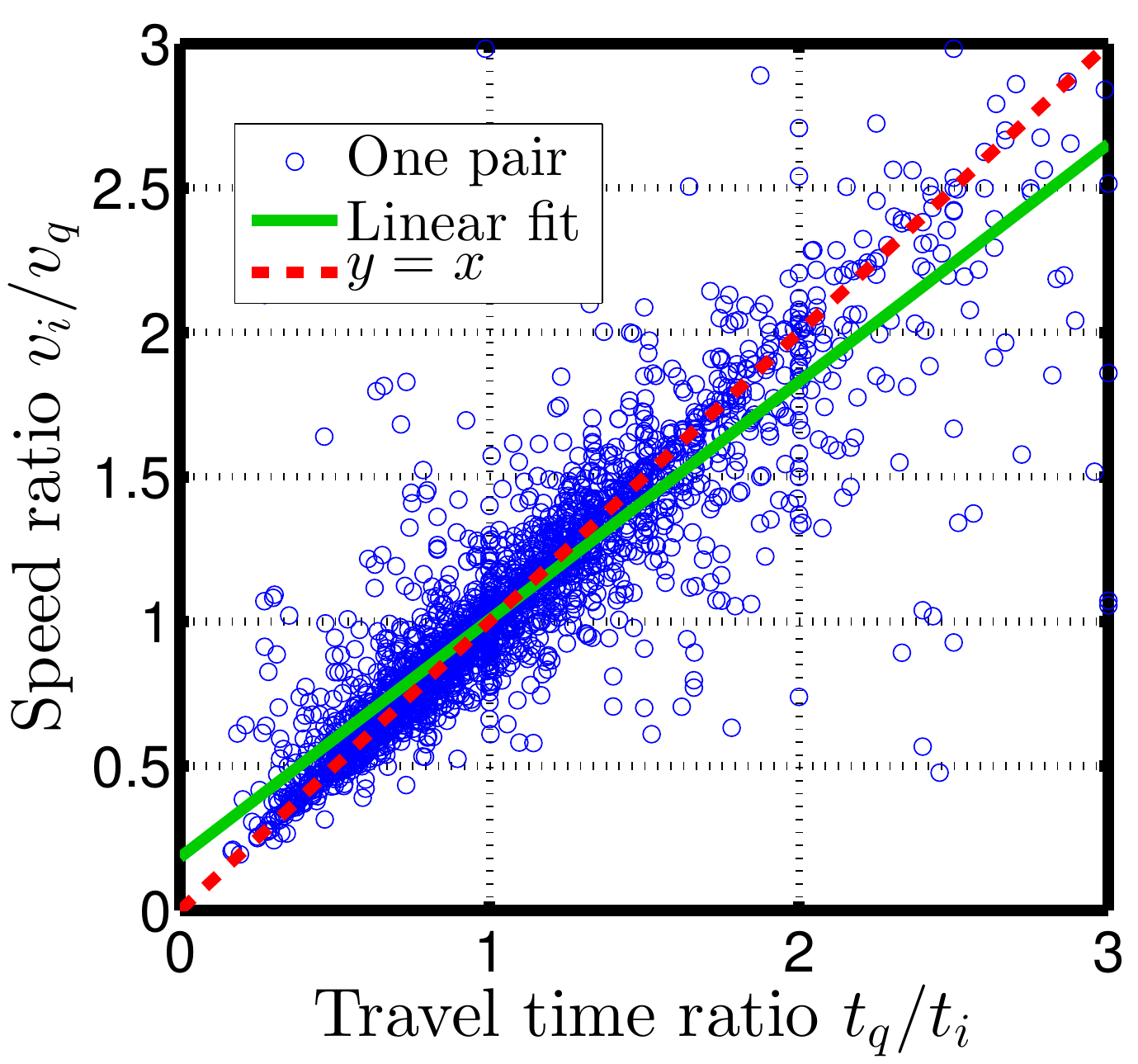}}
\subfigure[$\frac{v_i}{v_q} \approx \frac{V(s_i)}{V(s_q)}$\label{fig:assump2}]{\includegraphics[width=0.23\textwidth]{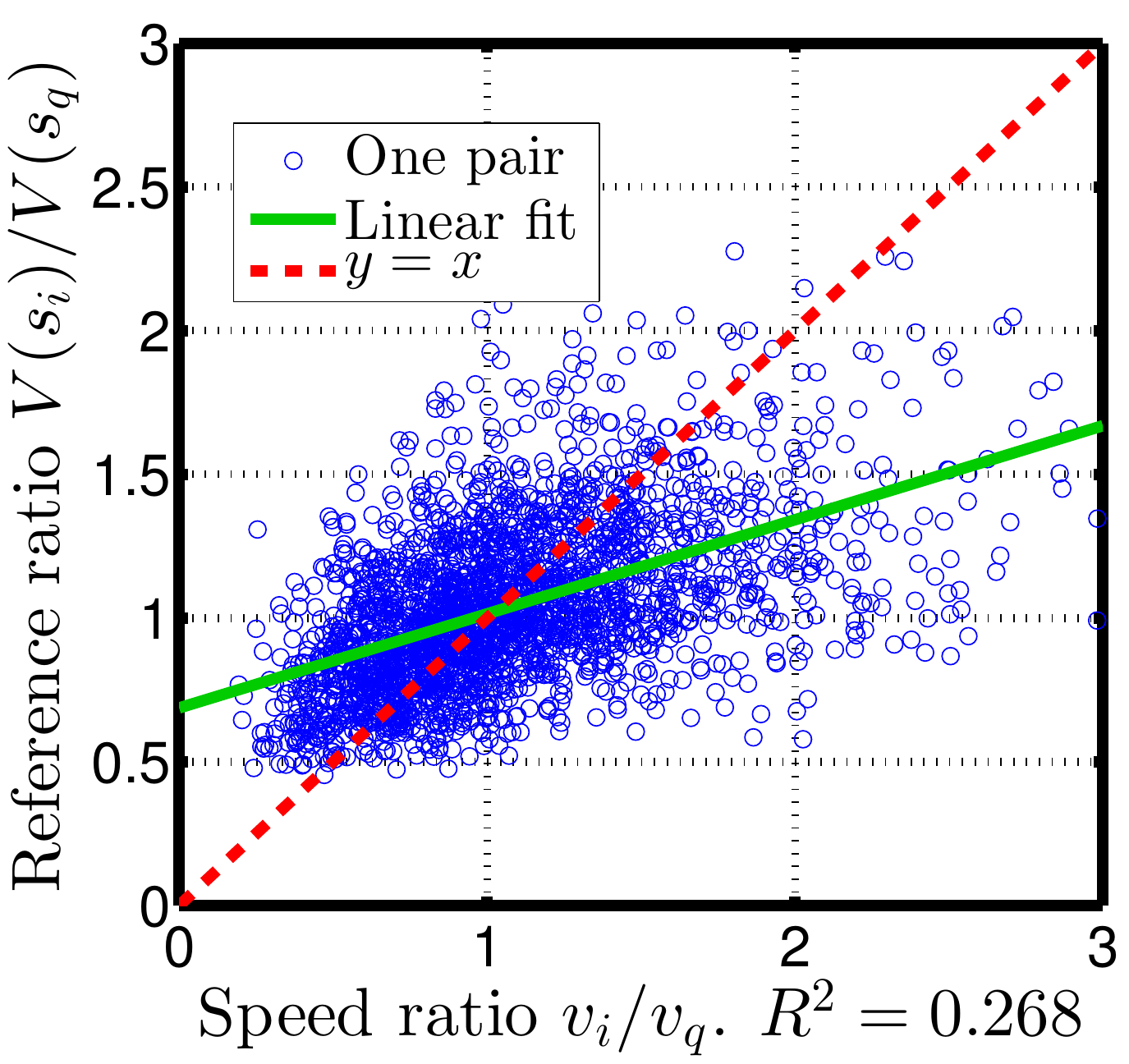}}
\caption{The validity of our assumptions. We randomly sample a set of neighboring trip pairs, and calculate the different ratios for each pair. We plot each trip pair as a blue point. The solid green line is the linear fit of all the points. The dotted red line is $y=x$. In (a), we can see that fitted line has a slope approximating $1$, which verifies our assumption $\forall \p_i \in \N(q), l_i \simeq l_q$. Similarly, in (b), we conclude that ratio of speed reference approximates the ratio of actual speed.}
\end{figure}

Now the question is, how can we derive a \emph{temporal scaling reference} to correspondingly adjust travel time on the neighboring trips.
We first define the \textbf{scaling factor} of a neighboring trip $\p_i$ on query trip $\q$ as:
\begin{equation}
s_i = \frac{t_q}{t_i}.
\end{equation}

One way to estimate $s_i$ is using the speed of $\p_i$ and $\q$. Let $v_i$ and $v_q$ be the speed of trip $\p_i$ and $\q$. Since we pick a small $\tau$ to extract neighboring trips of $\q$, it is safe to assume that $\forall \p_i \in \N(q)$, $l_i \simeq l_\q$. In Figure~\ref{fig:assump1}, on a sample set of neighboring trip pairs, we plot the inverse speed ratio against the travel time ratio. The solid line shows the actual relation between $\displaystyle\frac{t_q}{t_i}$ and $\displaystyle\frac{v_i}{v_q}$, which is very close to the dotted line $y=x$, which means that two ratios are approximately equivalent. With this assumption, we have:
\[
s_i = \frac{t_q}{t_i} = \frac{l_q / v_q}{l_i / v_i} \approx \frac{v_i}{v_q}.
\]

However, $v_q$ is unknown, so we need to estimate $\displaystyle\frac{v_i}{v_q}$. Since the average speed of all trips are stable and readily available, we try to build a bridge from the actual speed ratio to the corresponding average speed ratio for any given two trips. One solution is to assume the ratio between $v_q$ and $v_i$ approximately equals to the ratio between the average speed of all trips at $s_q$ and $s_i$. Formally, let $V(s)$ denote the average speed of all trips at timestamp $s$, we have an approximation of $s_i$ as
\begin{equation}
\label{eq:si}
s_i \approx \frac{v_i}{v_q} \approx \frac{V(s_i)}{V(s_q)}.
\end{equation}

This assumption is validated in Figure~\ref{fig:assump2}. For the sample set of neighboring trip pairs, the average speed ratio is plotted against actual speed ratio. Since the points are approximately distributed along the line $y=x$, we conclude that average speed ratio is a feasible approximation of the scaling factor.  We notice that the fitted line has a slope less than 1, which is mainly due to the anomaly in individual trips. Specifically, the speed of individual trips $v$ has a higher variance and some individual trip pairs have an extreme large ratio. On the other hand, the average speed $V(s)$ has a smaller variance and the ratio is confined to a more reasonable range $[0.5, 2]$.

\begin{figure}[t]
\centering
\subfigure[]{\includegraphics[width=0.234\textwidth]{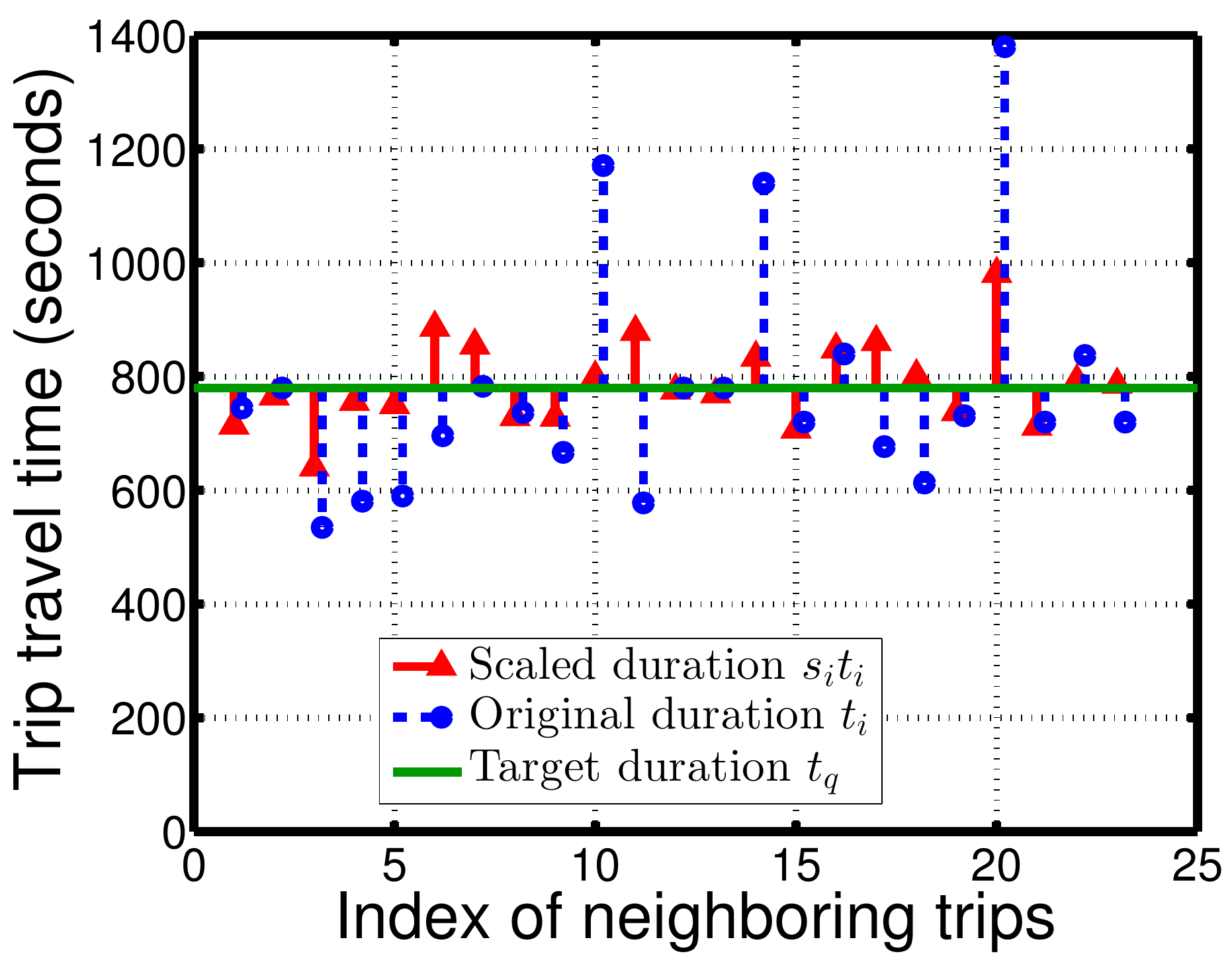}}
\subfigure[]{\includegraphics[width=0.234\textwidth]{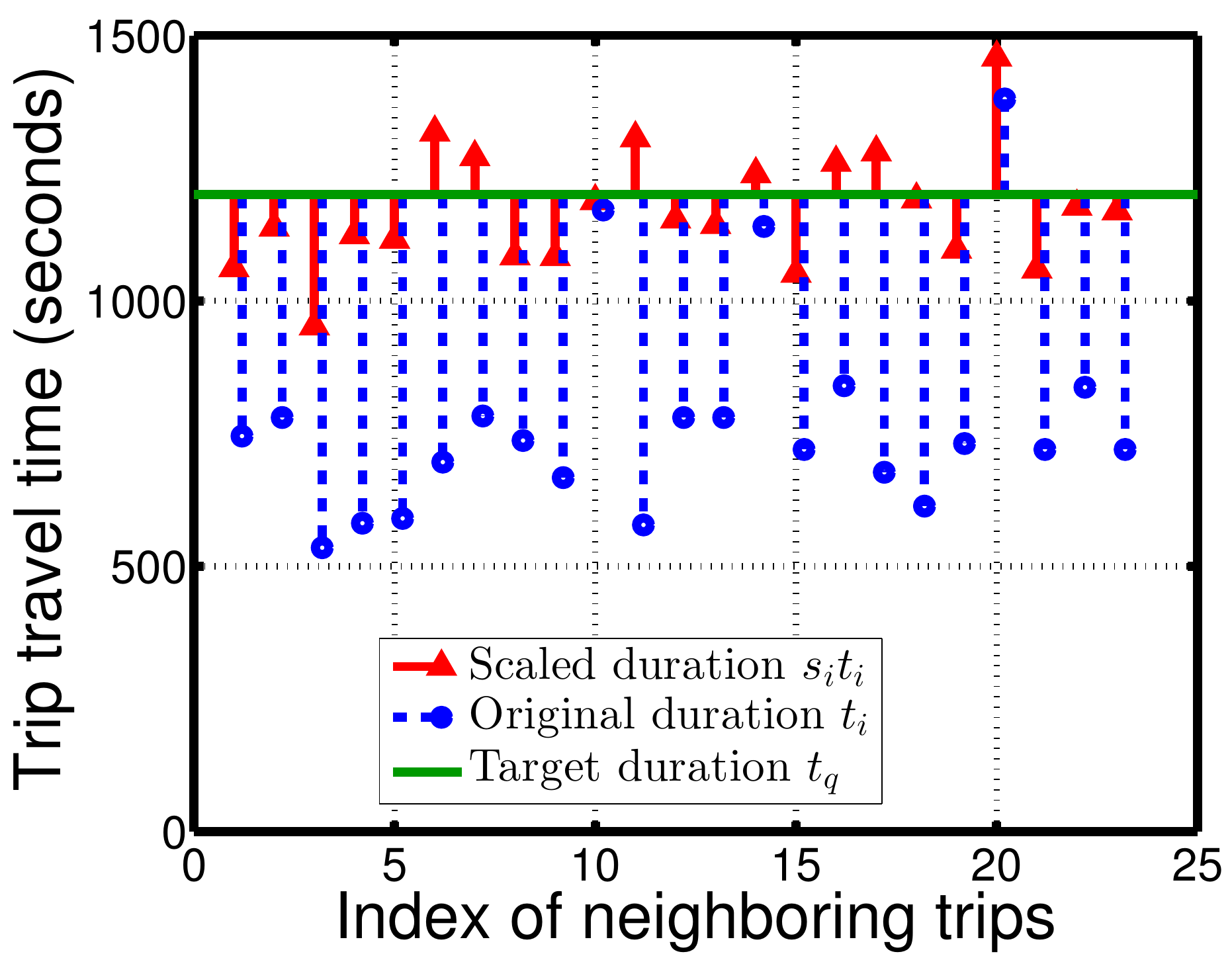}}
\caption{Two specific cases demonstrate the effectiveness of the scaling factor. The actual travel time $t_i$ of neighboring trips is plotted as blue circle, while the scaled travel time $s_it_i$ is plotted as red triangle. The horizontal green line shows the travel time of target trip $\q$. In (a), the actual travel times of historical trips have very large variance, because they are from different time slots with distinct traffic conditions. The scaling factor successfully reduces the variance in the actual travel time $t_i$. In (b), the target trip $\q$ occurred in rush hour, and thus $t_q$ is much longer than most of its neighboring trips. The scaling factor successfully fill in the gap between historical trips and $\q$.}
\label{fig:scale-cases}
\end{figure}

Next, we show the effectiveness of the scaling factor calculated from Equation~\ref{eq:si}. In Figure~\ref{fig:scale-cases}, we present two specific trips to demonstrate the intuition of how the scaling factor works. For each trip, we retrieve its historical neighboring trips, and then plot the actual travel time in circle together with the scaled travel time in triangle. From the Figure~\ref{fig:scale-cases}(a), we can see that the scaling factor helps reduce the variance in the travel time of neighboring trips. In Figure~\ref{fig:scale-cases}(b), trip $\q$ happened in rush hour, which takes longer time than most of its neighboring trips. This gap is successfully filled in by the scaling factor.

Considering such scaling factors using temporal speeds as the reference,  we can estimate the travel time of $\q$ using the neighboring trips as follows:
\begin{equation}
\widehat{t}_q =\frac{1}{|\N(\q)|} \sum_{\p_i \in \N(\q)} t_i \cdot \frac{V(s_i)}{V(s_q)}.
\end{equation}

\begin{figure}[t]
\centering
\includegraphics[width=0.35\textwidth]{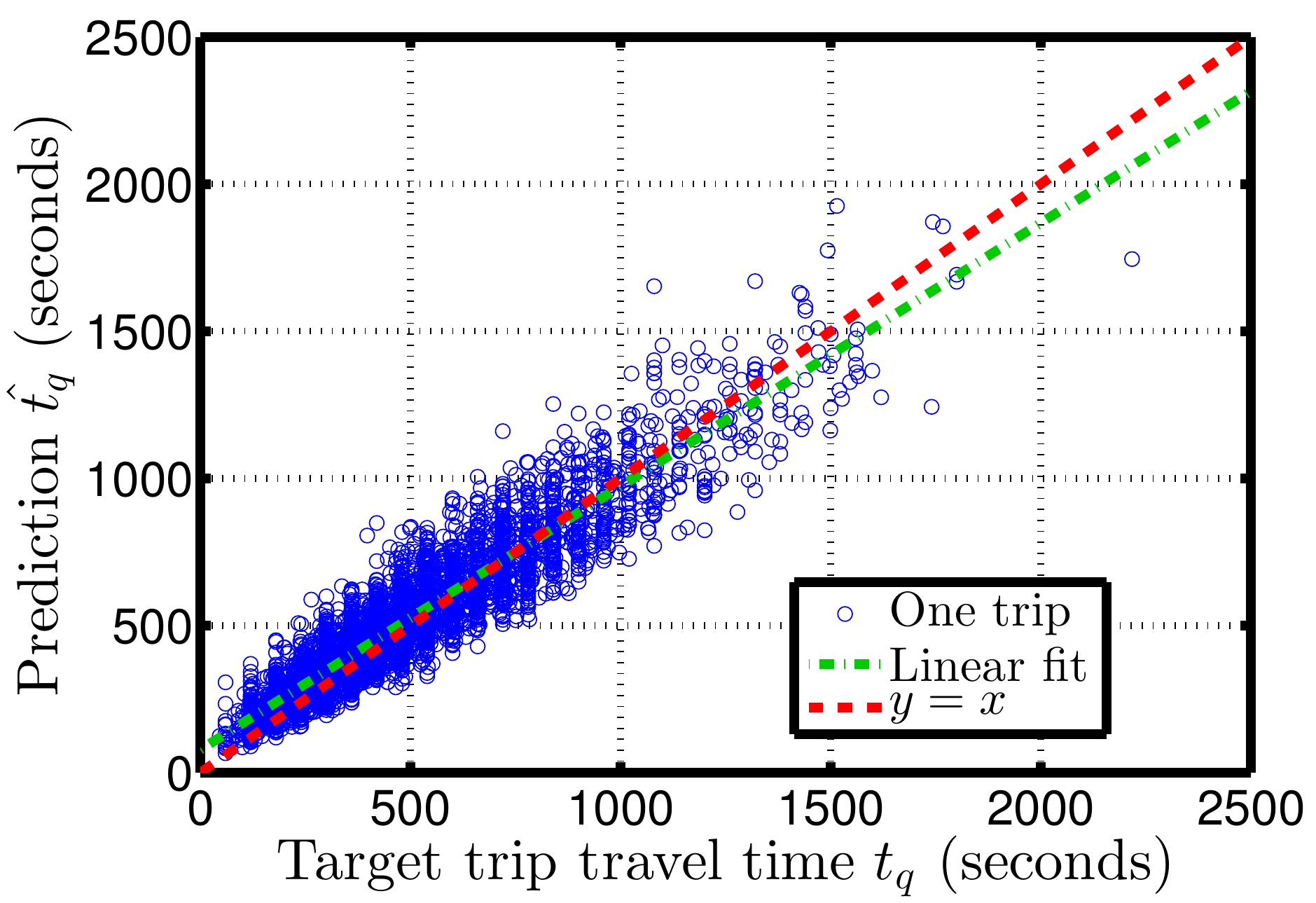}
\caption{Estimated travel time against actual travel time. Each point is a trip, where the estimation  $\hat{t_q}$ is plotted against $t_q$. The green line is the linear fitting of the points, which is closer to the red line $y=x$. This means the prediction is close to the actual value.}
\label{fig:estimator}
\end{figure}

We show the effectiveness of this predictor in Figure~\ref{fig:estimator}. Each point in the figure is a target trip. The prediction is plotted against the actual trip travel time. We can see that the prediction is close to the actual value. This indirectly implies the validity of assumptions we made previously.

In order to compute the average speed $V(s_i)$, we need to collect all the trips in $\D$ which started at time $s_i$. However, for the query trip $\q$, the starting time $s_q$ may be the current time or some time in the future. Therefore, no trips in $\D$ have the same starting time as $\q$. In the following, we discuss two approaches to predict $V(s_q)$ using the available data.

\subsection{Relative Temporal Speed Reference}
\label{sec:method:period}


In this section, we assume that  $V(s)$ exhibits a regular  daily or weekly pattern. We fold the time into a relative time window $\Tr = \{1, 2, \cdots, T\}$, where $T$ is the assumed periodicity. For example, using a weekly pattern with 1 hour as the basic unit, we have $T=7\times 24=168$. Using this relative time window, we represent the average speed of the $k$-th time slot as $V_k, \forall k \in \Tr$. We call $\{V_k| k \in \Tr\}$ \emph{relative temporal reference}. 

We use $k_i$ to denote the time slot to which $s_i$ belongs. As a result, we can write $V(s_i) = V_{k_i}$. To compute $V_{k_i}$, we collect all the trips in $\D$ which fall into the same time slot as $\p_i$ and denote the set as $S(\p_i)$. Then, we have 
\begin{equation}
V_{k_i} = \frac{1}{|S(\p_i)|} \sum_{\p_j \in S(\p_i)} \frac{l_j}{t_j}
\end{equation}

In Figure~\ref{fig:method:speed}, we present the weekly relative speed reference on all trips. We can see all the weekdays share similar patterns the rush hour started from 8:00 in the morning. Meanwhile, during the weekends, the traffic is not as much as usual during 8:00 in the morning.

The relative speed reference mainly has the following two advantages. First, the relative speed reference is able to alleviate the data sparsity issue. By folding the data into a relative window, we will have more trips to estimate an average speed with a higher confidence. Second, the computation overhead of relative speed reference is small, and we could do it offline.

\subsection{Absolute Temporal Speed Reference}
\label{sec:method:arima}

\begin{figure}[t]
\centering
\includegraphics[width=0.48\textwidth]{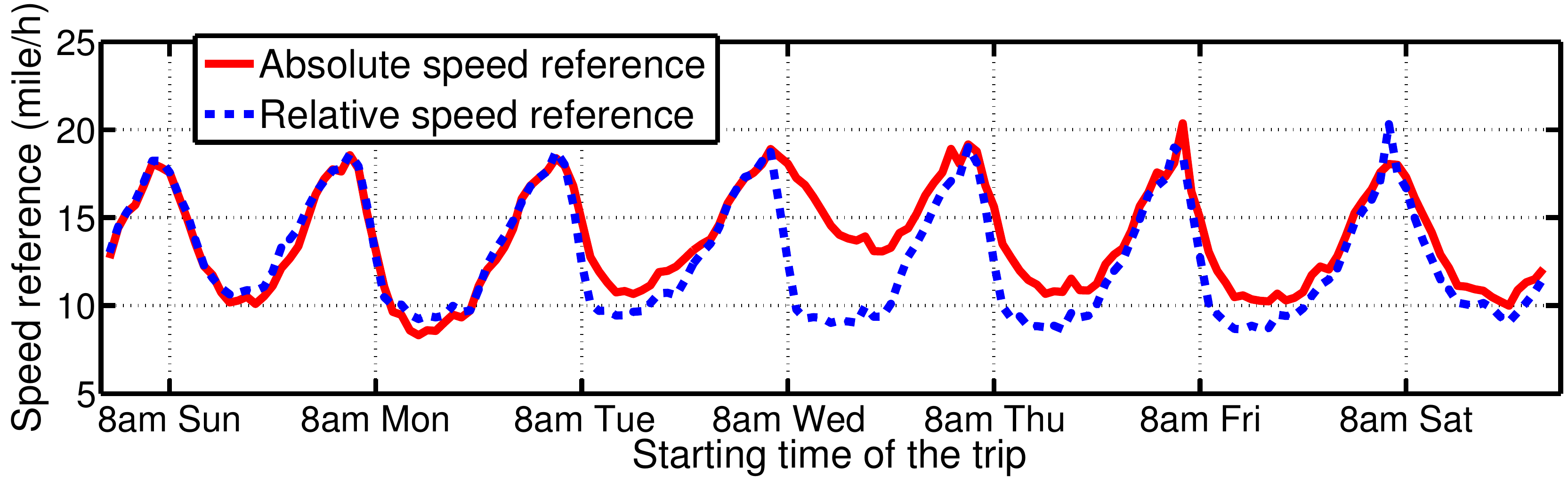}
\caption{Comparison between the absolute speed reference and the relative speed reference in the Christmas week (Dec 22, 2015 -- Dec 28, 2013). We can see the traffic condition is much better during Wednesday daytime than usual, because people are celebrating Christmas.}
\label{fig:comp-rela-abs}
\end{figure}

In the previous section, the relative speed reference assumes the speed follows daily or weekly regularity. However, in real scenario, there are always irregularities in the traffic condition. For example, during national holidays the traffic condition will significantly deviate from the usual days. In Figure~\ref{fig:comp-rela-abs}, we show the actual average traffic speed during the Christmas week (Dec 22, 2013 -- Dec 28, 2013). Compared with the relative speed reference, we can see that on Dec 25th, 2013, the traffic condition is better than usual during the day, since people are celebrating Christmas. Therefore, assuming we have enough data, it would be more accurate if we could directly infer the average speed at any time slot $t$ from the historical data.

In this section, we propose an alternative approach to directly capture the traffic condition at different time slots. We  extend our relative reference from Section~\ref{sec:method:period} to an \emph{absolute temporal speed reference}. To this end, we partition the original timeline into time slots based on a certain time interval (i.e., 1 hour). All historical trips are mapped to the absolute time slots $\Ta = \{ 1, 2, 3, \cdots\}$ accordingly, and the average speed $\{V_k |  k \in \Ta\}$ are calculated as the absolute temporal speed reference.

The challenge in absolute temporal speed reference is that for a given query trip with starting time $s_q$ in the near future (e.g. next hour), we need to estimate the speed reference $V(s_q)$.
We  estimate $V(s_q)$  by taking into account factors such as the average speed of previous hours, seasonality, and random noise. Formally, given the time series of average speed: $\{V_1, V_2, \ldots, V_{M}\}$, our goal is to compute $V_{M+1}$ as follows:
\begin{equation}
V_{M+1} = f(V_1, \ldots, V_M). 
\end{equation}

To tackle this problem, we adopt the ARIMA model for time series forecasting, and show how it can be used to model the seasonal difference (\ie, the deviation from the periodic pattern) for a time series.

\subsubsection{Overview of the ARIMA Model}

In statistical analysis of time series, the autoregressive integrated moving average (ARIMA) model~\cite{HyAt14} is a popular tool for understanding and predicting future values in a time series. Mathematically, for any time series $\{X_t\}$, let $L$ be the \emph{lag operator}:
\begin{equation}
LX_t = X_{t-1}, \forall t > 1.
\end{equation}
Then, the ARIMA$(p,d,q)$ model is given by
\begin{equation}
\left( 1-\sum_{i=1}^p \phi_i L^i\right) \left(1-L\right)^d X_t = \left(1+\sum_{i=1}^q \theta_i L^i\right) \epsilon_t,
\end{equation}
where parameters $p$, $d$ and $q$ are non-negative integers that refer to the order of the autoregressive, integrated, and moving average parts of the model, respectively. In addition, $\epsilon_t$ is a white noise process. 

In practice, the autocorrelation function (ACF) and partial autocorrelation function (APCF) is frequently used to estimate the order parameters $(p,d,q)$ from a time series of observations $\{X_1, X_2, \ldots, X_M\}$. Then, the coefficients $\{\phi_i\}_{i=1}^p$ and $\{\theta_i\}_{i=1}^q$ of the ARIMA$(p,d,q)$ can be learned using standard statistical methods such as the least squares.

\subsubsection{Incorporating the Seasonality}

In our problem, the average speed $V_t$ exhibits a strong weekly  pattern. Thus, instead of directly applying the ARIMA model to $\{V_t\}$, we first compute the sequence of seasonal difference $\{Y_t\}$:
\begin{equation}
Y_t = V_t - V_{t-T},
\label{eq:YV}
\end{equation}
where $T$ is the period (\eg, one week). Then, we apply the ARIMA model to $\{Y_t\}$:
\begin{equation}
\left( 1-\sum_{i=1}^p \phi_i L^i\right) \left(1-L\right)^d Y_t = \left(1+\sum_{i=1}^q \theta_i L^i\right) \epsilon_t.
\end{equation}
Note that our ARIMA model with the seasonal difference is a special case of the more general class of Seasonal ARIMA (SARIMA) model for time series analysis. We refer interested readers to~\cite{HyAt14} for detailed discussion about the model.

Suppose we use first order difference of $Y_t$, namely $d = 1$ and $(1-L)^d Y_t = Y_t - Y_{t-1}$. Then we have
\begin{equation}
Y_t = Y_{t-1} +  \sum_{i=1}^p \phi_i L^i (Y_t - Y_{t-1}) + \sum_{i=1}^q \theta_i L^i \epsilon_t + \epsilon_t
\label{eq:Y_t}
\end{equation}
Since the last term $\epsilon_t$ in Equation~(\ref{eq:Y_t}) is white noise, whose value is unknown but the expectation $E(\epsilon_t) = 0$, we have estimator $\hat{Y_t} = Y_t - \epsilon_t$. Together with Equation~(\ref{eq:YV}), we have
\begin{equation}
\hat{V_t} = \hat{Y_t} + V_{t-T}
\end{equation}

\subsection{The Effect of Geographic Regions}
\label{sec:method:neighborhoods}
\begin{figure}[h!]
\centering
\subfigure[Manhattan]{\includegraphics[width=0.23\textwidth]{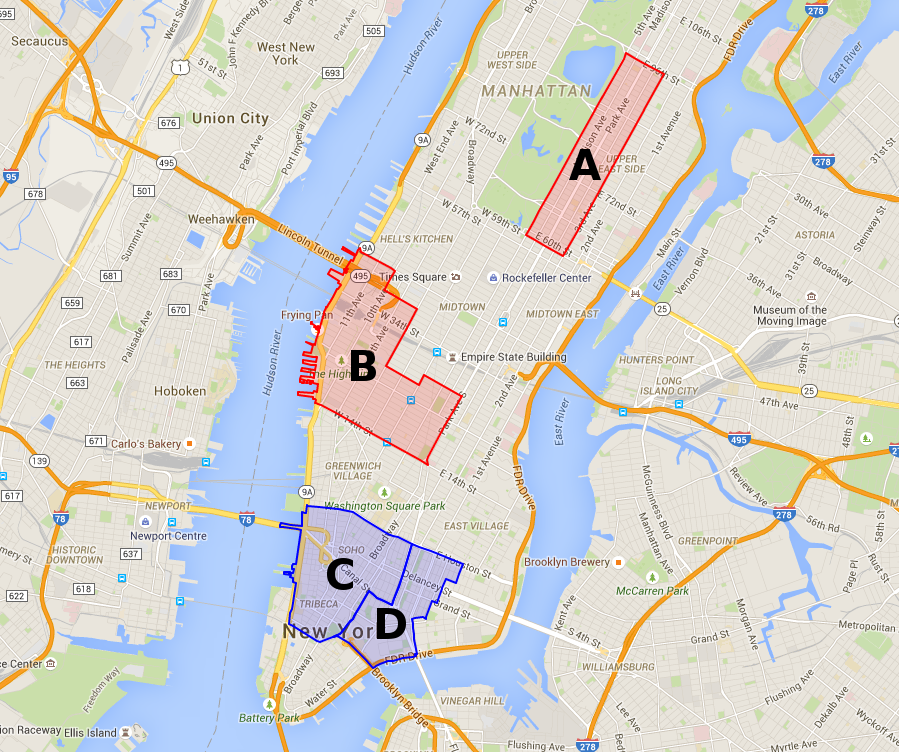}}
\subfigure[Speed reference]{\includegraphics[width=0.23\textwidth]{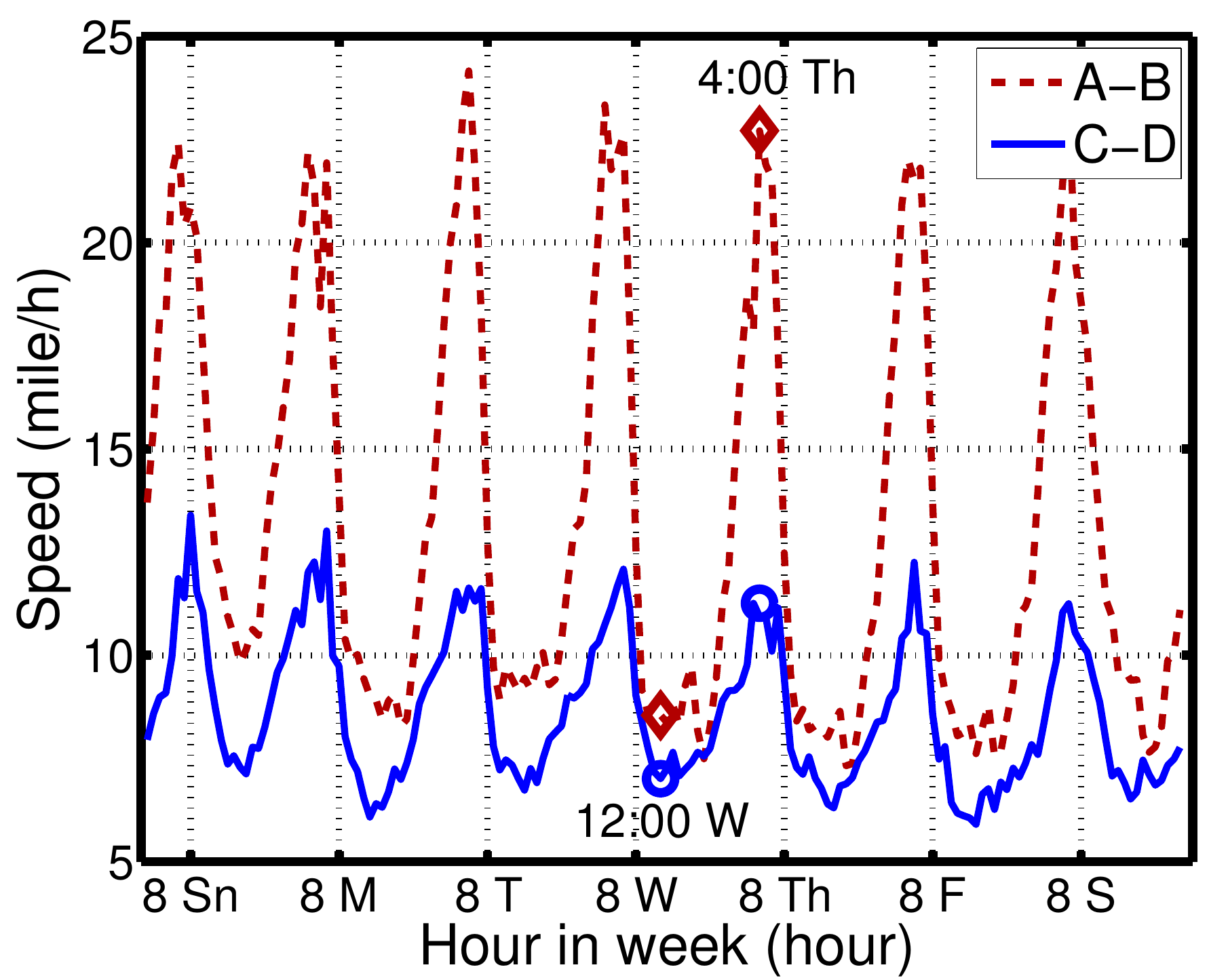}}
\caption[caption for figure]{Different region pairs have different traffic patterns. \footnotemark[1]}\label{fig:speed-ref}
\end{figure}

So far, we have assumed that the traffic condition follows the same temporal pattern across all geographic locations.
However, in practice, trips within a large geographic area (\eg, New York City) may have different traffic patterns, depending on the spatial locations. For example, in Figure~\ref{fig:speed-ref} we show the speed references of two different pairs of regions. The speed reference in region pairs ($A, B$) has a larger variation than ($C, D$). In particular, for ($A,B$) pair, the average speed is 22.7mph at 4:00 a.m. on Thursday and 8.5mph at 12:00 p.m. on Wednesday. But for ($C,D$) pair, the average speed is only 11.2mph and 7.0mph at these two corresponding times. The reason could be that $A$ is a residential area, whereas $B$ is a business district. Therefore, the traffic pattern between $A$ and $B$ exhibits a very strong daily peak-hour pattern. On the other hand, regions $C$ and $D$ are popular tourist areas, the speeds are constantly slower compared with that of ($A, B$). 

\footnotetext[1]{Figure~\ref{fig:speed-ref}(a) is generated with Google MAP API.}


The real case above suggests that refining the temporal references for different spatial regions could help us further improve the travel time estimation.  Therefore, we propose to divide the map into a set of $K$ neighborhoods $\{R_1, R_2, \ldots, R_K\}$. For each pair of neighborhoods $(R_i,R_j), 1\leq i,j \leq K$, we use $V_{i\rightarrow j}(s)$ to denote the average speed of all trips from $R_i$ to $R_j$ at starting time $s$. Let $\D_{i\rightarrow j}$ be the subset of trips in $\D$ whose origin and destination fall in $R_i$ and $R_j$, respectively. Then, $V_{i\rightarrow j}(s)$ can be estimated using $\D_{i\rightarrow j}$ in the same way as described in Sections~\ref{sec:method:period} and~\ref{sec:method:arima}.

\begin{figure}[t]
\centering
\includegraphics[width=0.23\textwidth]{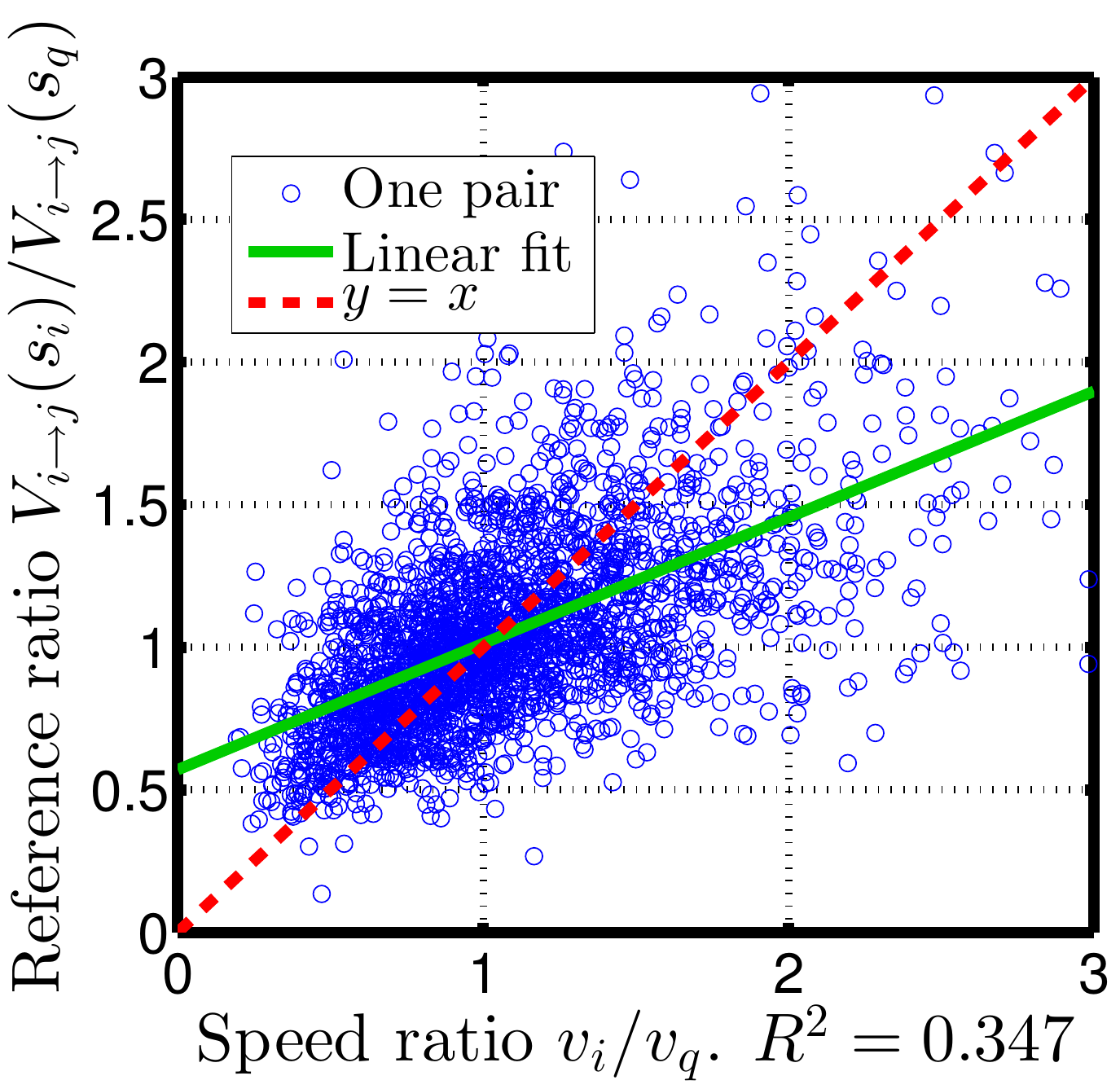}
\caption{Refining temporal reference with regions improves the performance. $\frac{V_{i\rightarrow j}(s_i)}{V_{i\rightarrow j}(s_q)} \approx \frac{v_i}{v_q}$. Compared with Figure~\ref{fig:assump2}, the slope of linear fit is closer to 1 and the $R^2$ is larger, which means the region-based temporal reference approximates the actual speed ratio better.}
\label{fig:region-ref}
\end{figure}

To show the effectiveness of refining the temporal reference for each region pairs. 
In Figure~\ref{fig:region-ref}, we plot the refined temporal reference ratio against the actual speed ratio of a pair of trips. Comparing with the Figure~\ref{fig:assump2}, we see that the linear fit has a slope closer to $1$, which means the two ratios are closer, and the $R^2$ is larger, which means the line fits better.

\subsection{Time Complexity Analysis}

Our approach has three steps: 1) mapping training trips into grids; 2) extracting neighbors of a given OD pair; and 3) estimating travel time based on the neighbors. 

\textbf{Step 1.} In order to quickly retrieve the neighboring trips, we employ a raster partition of the city (e.g., 50 meters by 50 meters grid in our experiment), and preparing $N$ training trips takes $O(N)$ time. This step can be preprocessed offline. 

\textbf{Step 2.} Given a testing trip $\q$, we find its corresponding origin gird in $O(1)$ time. Then, retrieving all the neighboring trips in the same grid would take $O(N/h)$ time if the trips are uniformly distributed, where $h$ is the total number of grids. In practice, however, the number of trips in each grid follows a long tail distribution. Therefore, the worst case time complexity of retrieving neighboring trips would be $O(\alpha\cdot N)$, where $\alpha\cdot N$ is the number of trips in the most dense grid ($\alpha = 0.01$ in NYC taxi data).  

\textbf{Step 3.} After retrieving the neighboring trips, the time complexity of calculating travel time of $\q$ is  $O(|\N(\q)|)$, where $|\N(\q)|$ is the number of neighbors of trip $\q$. The temporal speed references can be computed offline. The relative speed reference takes $O(N)$ time, and  ARIMA model can be trained  in $O(M^2)$ time, where $M$ is the number of time slots. During the online estimation, looking-up relative speed reference takes $O(1)$ time, whereas using ARIMA to predict the current speed reference takes $O((p+q)\cdot d)$ time, where $p$, $q$, and $d$ are the parameters learned during model training ($p=2$, $q=0$, and $d=1$ in our experiment).

Therefore, the time complexity of online computation is $O(\alpha\cdot N)$ in the worst case. In order to serve a large amount of batch queries, we can use multi-threading to further boost the estimation time. 

\nop{
\subsection{Temporal Dynamics without Trip Distance}

When information about the distance for trips in $\D$ is not available, we discuss an alternative approach to estimating the travel time $t_q$. Let $T(s)$ denote the average travel time of all trips which start at time $s$, we make the following assumption.
\begin{assumption}
For any trip $\p_i\in \N(\q)$, the ratio of $t_i$ and $t_q$ is equal to the ratio of the average travel time of all trips at time $s_i$ and $s_q$:
\begin{equation}
\frac{t_i}{t_q} = \frac{T(s_i)}{T(s_q)}.
\end{equation}
\label{ass:time}
\end{assumption}

With this assumption, the travel time of $\q$ can be estimated as:
\begin{equation}
\widehat{t}_q = \frac{1}{|\N(\q)|} \sum_{\p_i \in \N(\q)} t_i \cdot \frac{T(s_q)}{T(s_i)}.
\end{equation}
Note that $T(s)$ can be estimated via periodicity or a seasonal ARIMA model in the same way as $V(s)$. In addition, following the discussion in Section~\ref{sec:method:neighborhoods}, we can estimate an average travel time $T_{i\rightarrow j}(s)$ for each region pair $(R_i,R_j)$.
}

%% file: 3method-noise.tex

\section{Filtering Outliers}
\label{sec:noise}

In the NYC taxi data, we observe many anomalous trips, which cause large errors in our estimation. Malfunction of tracking device might be the main cause of outliers. For example, there are a large amount of trips with reported travel time as zero second, while their actual distances are non-zero. 

Some trip features are naturally correlated, such as distance vs. time and trip distance vs. distance between endpoints. If a trip significantly deviates from such correlation, it is very likely to be an outlier. Specifically, the correlation of distance and time could help us find anomalous trips with extremely fast or slow travel speeds; the correlation of trip distance and distance between endpoints could help us find detour trips. In principle, any outlier detection algorithm can be used to clean the data. In this paper, we assume there exists a linear correlation between some feature pairs. Based on this assumption, we design a linear regression model to fit the data, and employ Maximize Likelihood Estimation to identify outliers.

Our linear data model is as follows. Suppose we are interested in capturing the linear correlation between feature $X$ and feature $Y$,  we have
\[ Y = \omega_1 X + \omega_2 +  \epsilon, \]
where $\omega_1$, $\omega_2$ are the linear coefficients, and $\epsilon$ is the fitting error. The error term $\epsilon$ has a distribution that is a mixture of a Gaussian distribution (with zero mean and unknown variance) with (unknown) probability $(1-p)$ and a t-distribution with probability $p$. We fit the model to the data using variational inference. For every data point, the inference procedure provides a number $t_i$ which is an estimate of the probability that trip $\p_i$ is an outlier. After learning $p$, we set the top $p\cdot N$ trips with largest $t_i$ values as outliers. Details of the outlier filtering algorithm are put into the supplementary material\footnote{\url{https://www.dropbox.com/s/c62ntpeplvxfjla/outlier.pdf?dl=0}} due to space constraints. As an example, if $Y$ is set to trip time and $X$ is set to trip distance, then this approach filters outlying trips with speed outside the range $[2.03 mph, 52.74 mph]$. 

\nop{
 Term $\epsilon$ is introduced to capture outliers. If a trip is normal, $\epsilon$ comes from a Gaussian distribution $f_G$; if a trip is an outlier,  $\epsilon$ follows a T-distribution, $f_T$, since the number of outliers is small and the variance is unknown.

To obtain the data likelihood, we add several additional variables. Suppose the portion of outliers is $p$, then $\epsilon$ follows $f_G$ with probability $1-p$ and follows $f_T$ with probability $p$. For each trip $\p_i$, we introduce an outlier measure $t_i$ taking value from $0$ to $1$, and $t_i = 1$ indicates $\p_i$ is an outlier. The complete data likelihood is given as

\begin{equation}
\label{eq:outliers-mle}
L(\mathbf{\omega}, \theta) = \prod_{i=1}^n \left[ (1-p) \cdot f_G \right]^{1-t_i} \times \left[ p \cdot f_T \right]^{t_i}
\end{equation}

We can identify outliers by solving the MLE problem with standard EM algorithm. After getting $t_i$ for all trips, we find the top $p$ trips with largest $t_i$ values as outliers. 

}

%% file: 4experiment.tex

\section{Experiment}
\label{sec:exp}

In this section, we present a comprehensive experimental study on two real datasets. All the experiments are conducted on a 3.4 GHz Intel Core i7 system with 16 GB memory. We have released our code and sample data via an anonymous Dropbox link\footnote{\url{https://www.dropbox.com/s/uq2w50kgmgzymv3/traveltime.tar.gz?dl=0?}}.

\subsection{Dataset}
We conduct experiments on datasets from two different countries to show the generality of our approach.

\subsubsection{NYC Taxi}
A large-scale New York City taxi dataset has been made public online~\cite{nyctaxi}. The dataset contains $173,179,771$ taxi trips from 2013/01/01 to 2013/12/31. Each trip contains information about pickup location and time, drop off location and time, trip distance, fare amount, etc. We use the subset of trips within the borough of Manhattan (the boundary is obtained from \url{wikimapia.org}), which has $132,766,605$ trips. After filtering the outliers, there are $\bf 127,534,711$ trips left for our experimental evaluation. On average, we have $349,410$ trips per day.

\begin{figure}[htb]
\centering
\subfigure[Spatial distribution]{\includegraphics[width=0.238\textwidth]{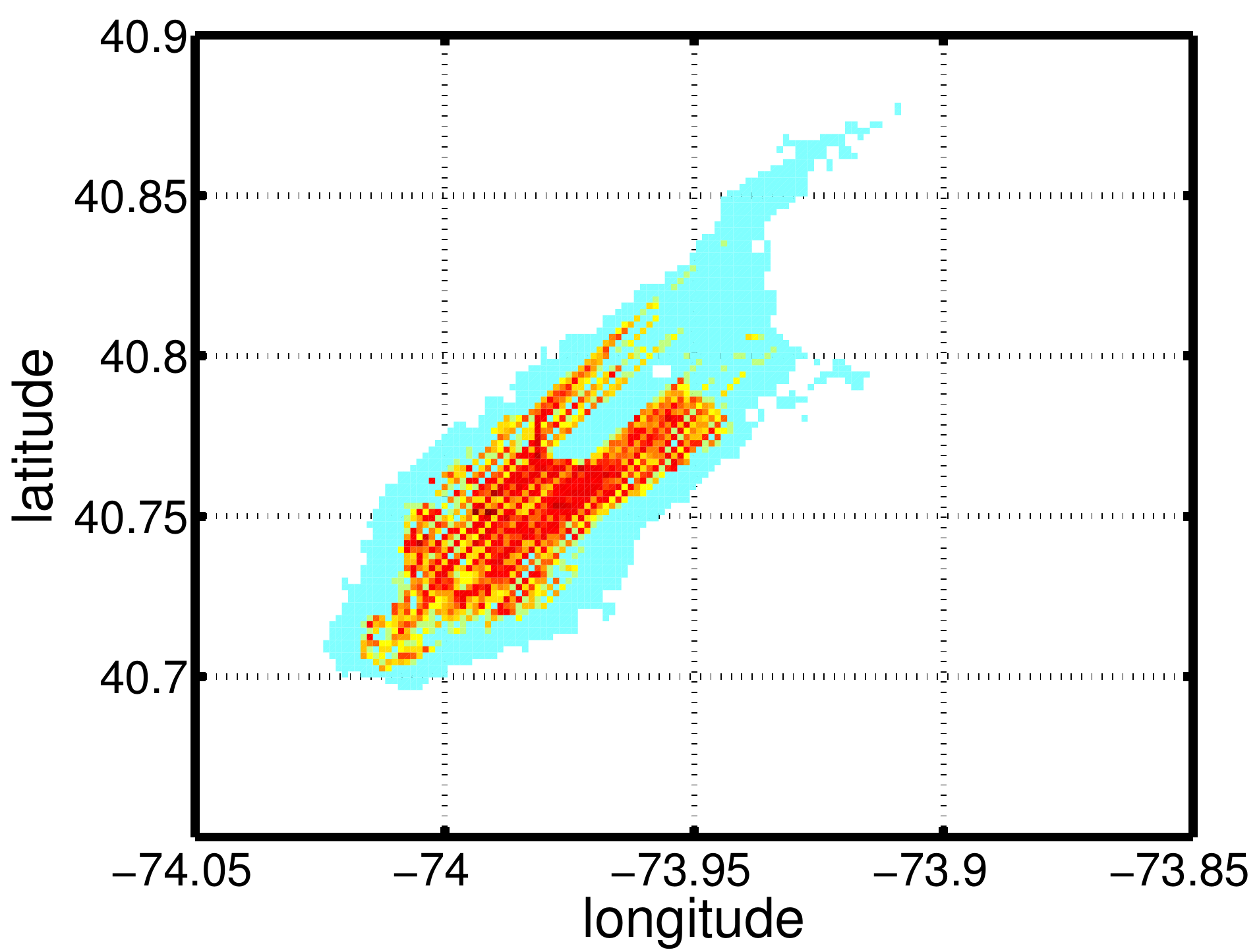}}
\subfigure[\# trips per day]{\includegraphics[width=0.23\textwidth]{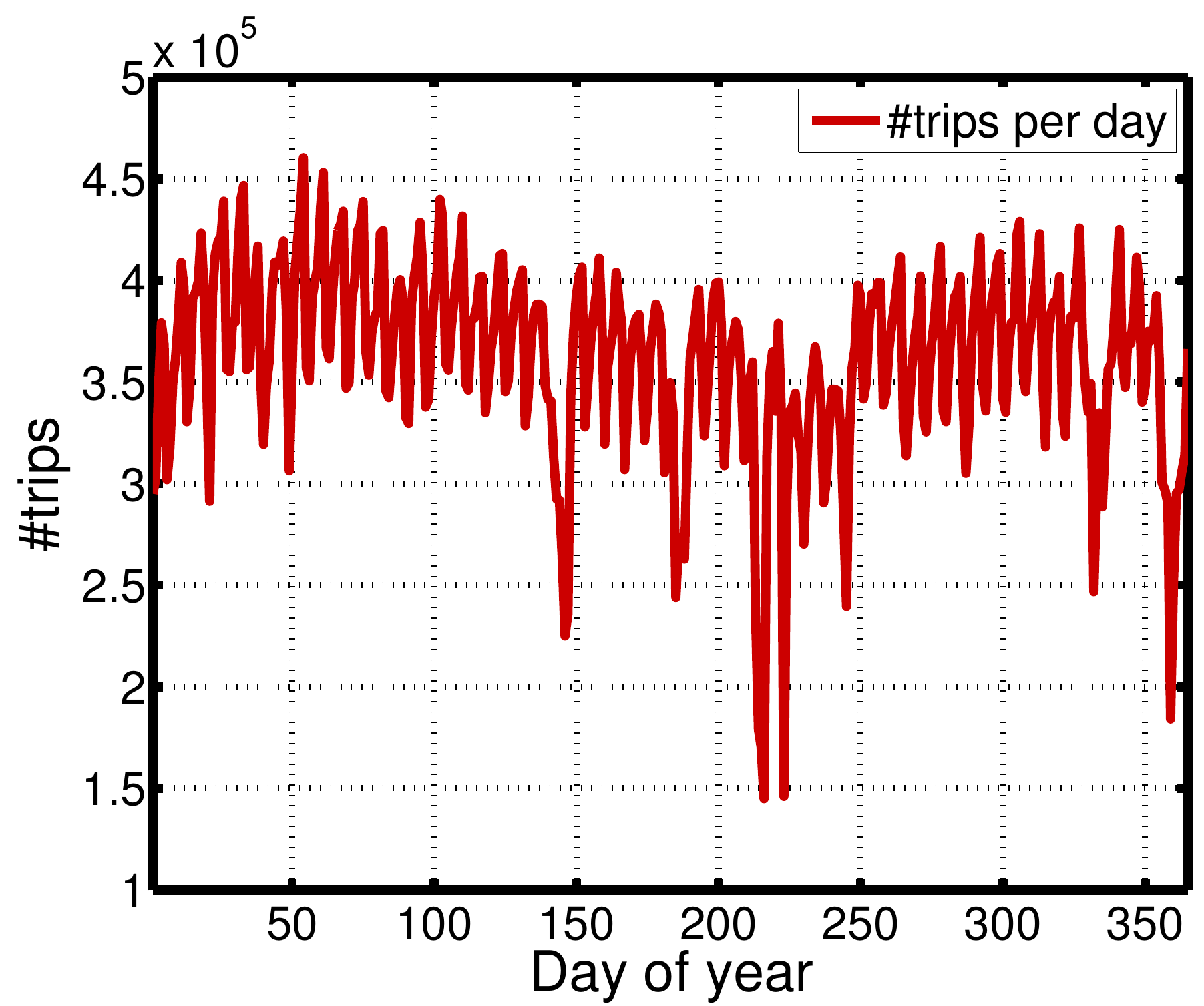}}
\subfigure[Trip time]{\includegraphics[width=0.23\textwidth]{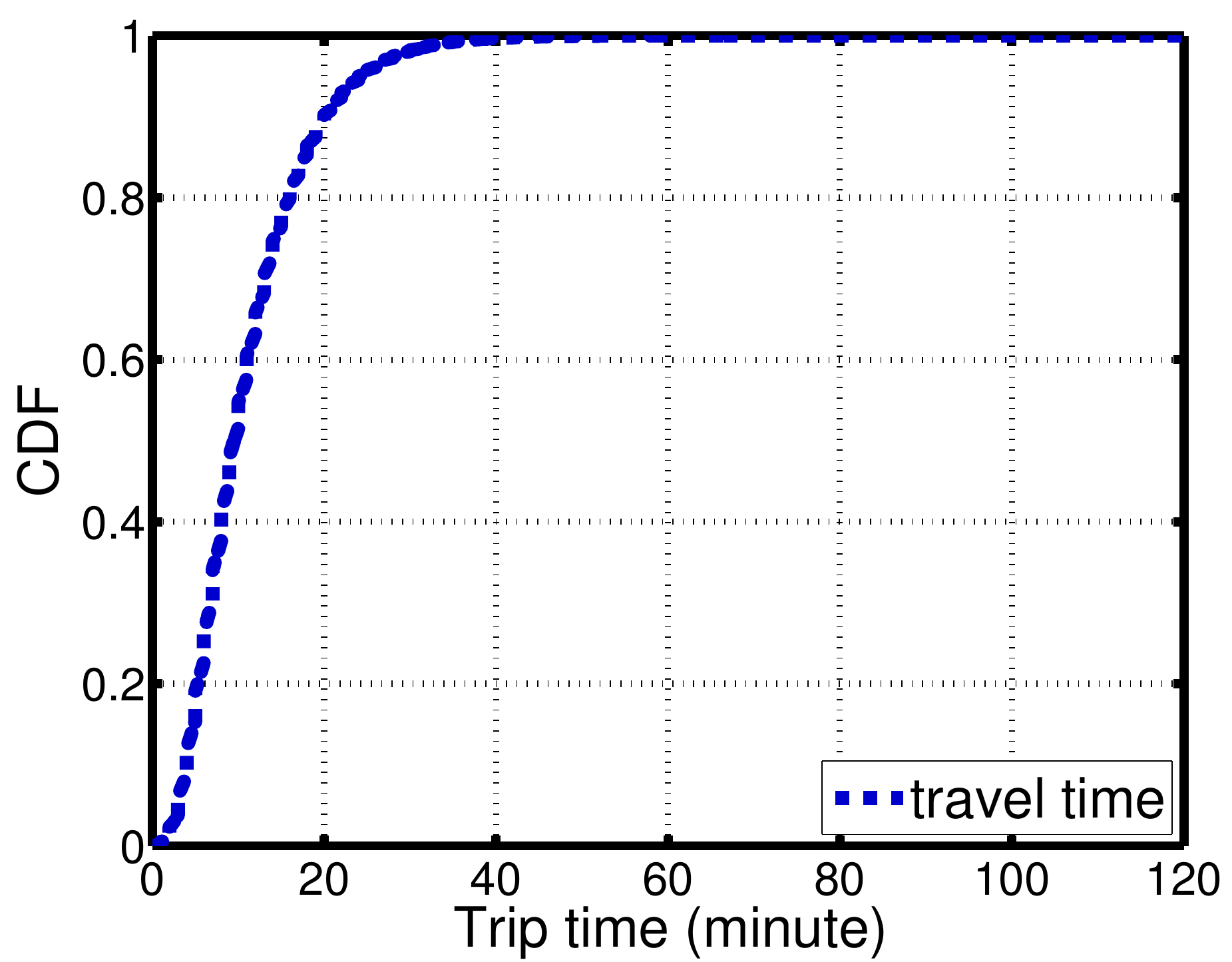}}
\subfigure[Trip distance]{\includegraphics[width=0.23\textwidth]{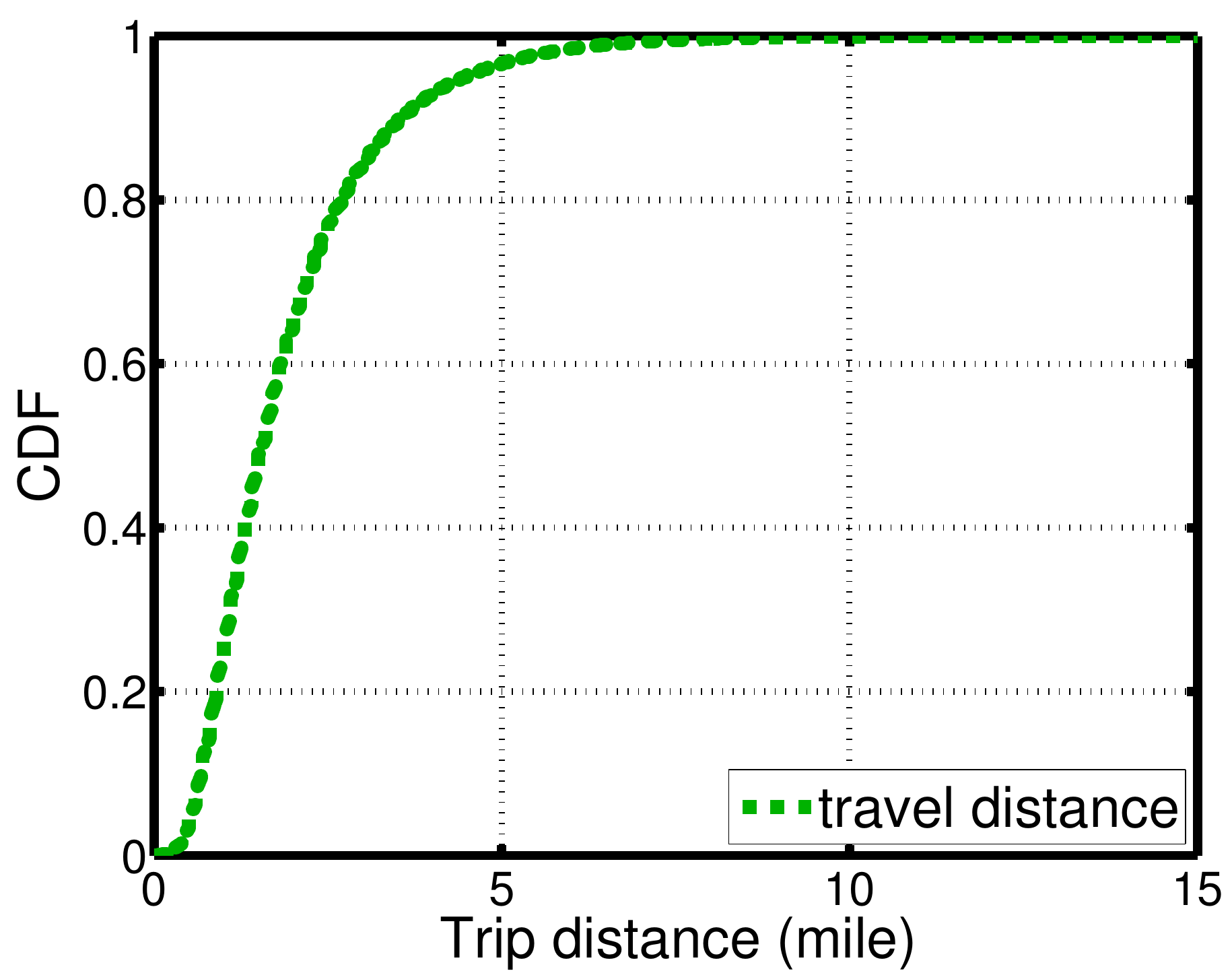}}
\caption{NYC data statistics. See text for explanations.}
\label{fig:data}
\end{figure}

Figure~\ref{fig:data}(a) shows the distribution of GPS points of the pickup and dropoff locations. Figure~\ref{fig:data}(b) shows the number of trips per day over the year 2013. Figure~\ref{fig:data}(c) and Figure~\ref{fig:data}(d) show the empirical CDF plot of the trip distance and trip time. About $56\%$ trips have trip time less than $10$ minutes, and about $99\%$ trips have trip time less than $30$ minutes. The mean and median trip time are $636$ (second) and $546$ (second). The mean and median trip distance are $1.935$ (mile) and $1.6$ (mile).

\subsubsection{Shanghai Taxi}
In order to compare with the existing route-based methods, we use the Shanghai taxi dataset with the trajectories of $2,600$ taxis during two months in 2006. A GPS record has following fields: vehicle ID, speed, longitude, latitude, occupancy, and timestamp. In total we have over $300$ million GPS records. We extract the geographical information of Shanghai road networks from OpenStreetMap.

To retrieve taxi trips in this dataset, we rely on the occupancy bit of GPS records. This occupancy bit is 1 if there are passengers on board, and 0 otherwise. For each taxi, we define a trip as consecutive GPS records with occupancy equal to 1. We get $5,815,470$ trips after processing the raw data. The distribution of trip travel time is similar to that of NYC taxi data in Figure~\ref{fig:data}(c). Specifically, about half of the trips have travel time less than $10$ minutes. 

\subsection{Evaluation Protocol}

\noindent \textbf{Methods for evaluation.} We systematically compare the following methods:
\begin{itemize}[leftmargin=*]
\item Linear regression (\lr). We train a linear regression model with travel time as a function of the L1 distance between pickup and drop off locations. This simple linear regression serves as a baseline for comparison.
\item Neighbor average (\avg). This method simply takes the average of travel times of all neighboring trips as the estimation.
\item Temporally weighted neighbors (\temp). This is our proposed method using temporal speed reference to assign weights on neighboring trips. We name the method using relative-time speed reference as \temprel (Section~\ref{sec:method:period}). And we call the method using ARIMA model to predict absolute temporal reference as \temparima (Section~\ref{sec:method:arima}).
\item Temporal speed reference by region (\tempR). This is our improved method based on \temp by considering the temporal reference for different region pairs (Section~\ref{sec:method:neighborhoods}).
\item Segment-based estimator (\seg). This method estimates the travel time of each road segment individually, and then aggregate them to get the estimation of a complete trip (a baseline method used in \cite{WZX14}).
\item Subpath-based estimator (\subpath). One drawback of \seg is that the transition time at intersections cannot be captured. Therefore, Wang et al. \cite{WZX14} propose to concatenate subpaths to estimate the target route, where each sub-path is consisted of multiple road segments. For each sub-path, \subpath estimates its travel time by searching all the trips that contain this sub-path.
\item Online map service (\bing and \baidu). We also compare our methods with online map services. We use Bing Maps~\cite{bingmap} for NYC taxi dataset and Baidu Maps~\cite{baidumap} for Shanghai dataset. We use Bing Maps instead of Google Maps for two reasons: (1) Bing Maps API allows query with current traffic for free whereas Google does not provide that; and (2) Bing Maps API allows 125K queries per key per year but Google only allows 2.5K per day. To consider traffic, we send queries to Bing Maps at the same time of the same day (in a weekly window) as the starting time of the testing trip. Due to national security concerns, the mapping of raw GPS data is restricted in China~\cite{chinaGeoRestrict}. So we use Baidu Maps instead of Bing Maps for Shanghai taxi dataset.
\end{itemize}

\noindent \textbf{Evaluation metrics.} Similar to \cite{WZX14}, we use mean absolute error (MAE) and mean relative error
(MRE) to evaluate the travel time estimation methods:
\[
MAE = \frac{\sum_i |y_i - \hat{y}_i|}{n}, MRE = \frac{\sum_i |y_i - \hat{y}_i|}{\sum_i y_i},
\]
where $\hat{y}_i$ is the travel time estimation of test trip $i$ and $y_i$ is
the ground truth. Since there are anomalous trips, we also use the median
absolute error (MedAE) and median relative error (MedRE) to evaluate the methods:
\[MedAE = median( |y_i - \hat{y}_i|), MedRE = median\left(\frac{|y_i - \hat{y}_i|}{y_i}\right),\]
where $median$ returns the median value of a vector.

\subsection{Parameter Setting}

Before conducting the comparison experiments, we first discuss how to set the parameter for
our proposed method. There is only one parameter in our proposed method, the
distance threshold $\tau$ to define the neighbors. We use the first $11$ months from NYC dataset to study the performance w.r.t. parameter $\tau$. For computational
efficiency, we partition the map into small grids of 50 meters by 50 meters. The
distance in Eq.~\eqref{eq:neighbor} is now defined as the L1 distance between two
grids. For example, if $p_1$ is a neighboring trip of $q$ for $\tau=0$, it means
that the pickup (and drop off) location of $p_1$ is in the same grid as the pickup
(and drop off) location of $q$. If $\tau=1$, a neighboring trip has endpoints in the same or adjacent grids.

\begin{figure}[htb]
\centering
\subfigure[Performance]{\includegraphics[width=0.23\textwidth]{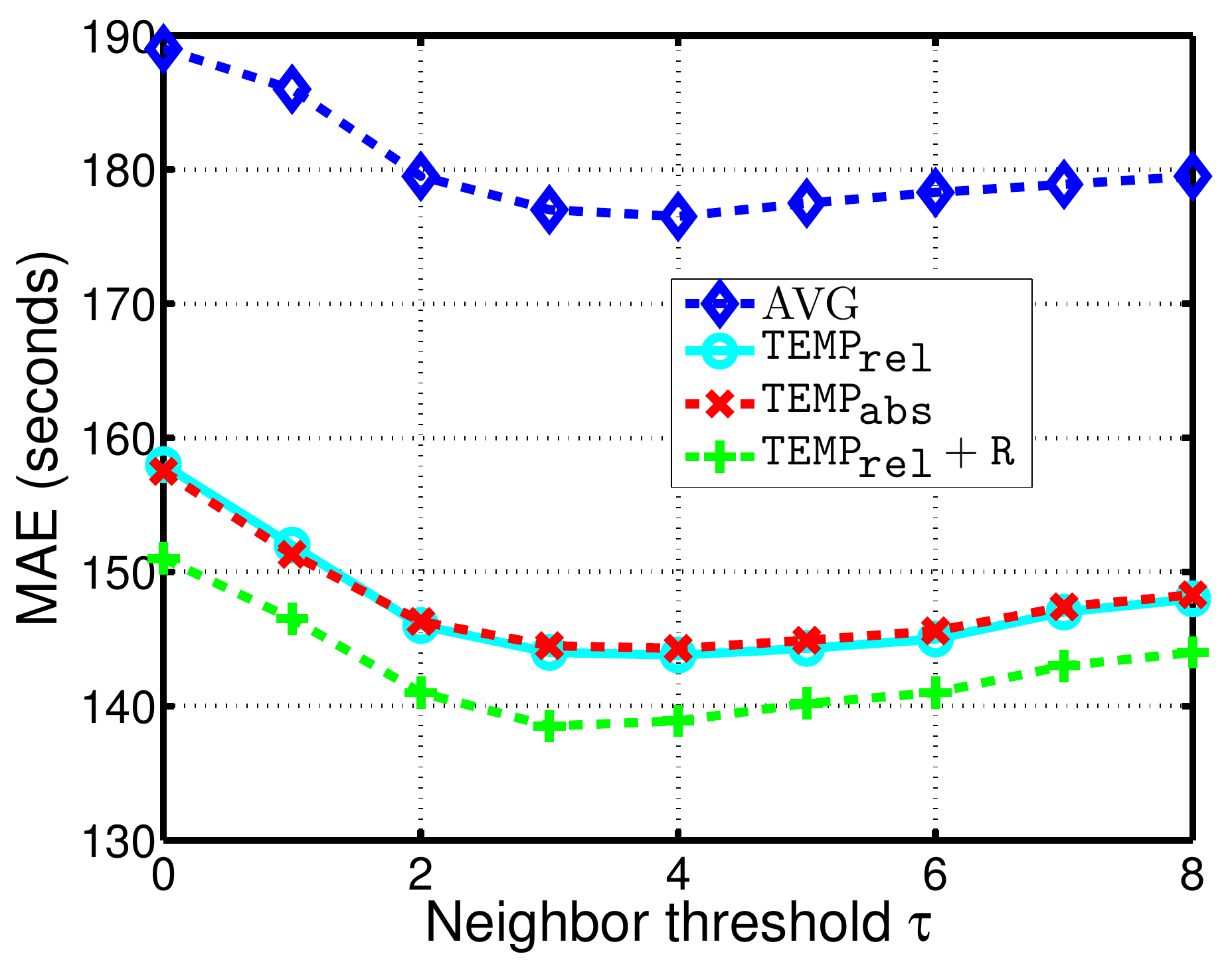}}
\subfigure[Coverage]{\includegraphics[width=0.23\textwidth]{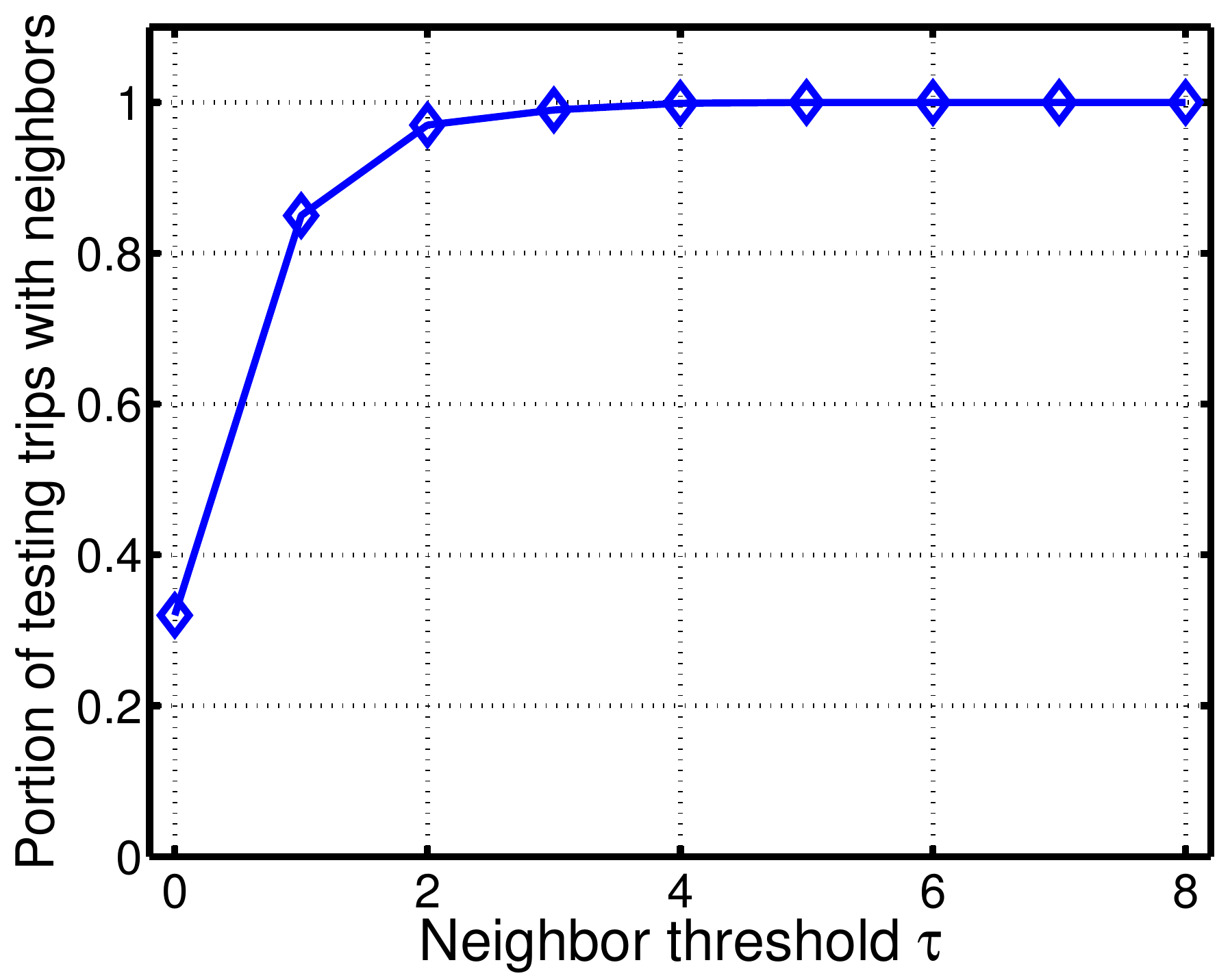}}
\caption{Estimation error and testing coverage w.r.t. neighbor threshold $\tau$.}
\label{fig:para}
\end{figure}

A larger $\tau$ will retrieve more neighboring trips for a testing trip and thus could give an estimation with a higher confidence. However, a larger $\tau$ also implies that the neighboring trips are less similar to the testing trip, which could introduce prediction errors. Therefore, it is crucial to identify the
optimal $\tau$ that balance estimation confidence and neighbors' similarity. In Figure~\ref{fig:para}(a), we plot the empirical relationship between the MAE and $\tau$. We can see that MAE is the lowest when $\tau$ is in the range of $[3,6]$, striking a good balance between confidence and similarity.
\nop{
  The reason is that, when $\tau$ is very small (less than $3$), the
number of neighboring trips is small, and the prediction is less reliable. On
the other hand, when $\tau> 6$, trips with completely different origin and
destination are identified as neighboring trips, which is inappropriate.}

We also want to point out that we need to have at least one neighboring trip in order to give an estimation of a testing trip. Obviously, the larger $\tau$ is, the more testing trips we can cover. Figure~\ref{fig:para}(b) plots the percentage of testing trips with at least one neighbor. In the figure is shows that if $\tau=0$, there are only $32\%$ of trips are predictable. When $\tau=3$, $99\%$ trips are predictable. Considering the both aspects, we find $\tau=3$ an appropriate setting for the following experiments.

\subsection{Performance on NYC Data}

\subsubsection{Overall Performance on NYC Data}

In our experiment, we use trips from the first 11 months (i.e., January to November) as training and the ones in the last month (i.e., December) as testing. As for the \bing method, due to the limited quota, we sample $260k$ trips in December as testing. The overall accuracy comparison are shown in Table~\ref{tab:comp-nyc}. Since NYC data only have endpoints of trips, we cannot compare our method with route-based methods. We will compare with theses methods using Shanghai data in Section~\ref{sec:exp:shanghai}.

\begin{table}[htdp]
\caption{Overall Performance on NYC Data}\label{tab:comp-nyc}
\begin{center}
\begin{tabular}{|l|l|l|l|l|}
\hline
Method & MAE (s) & MRE & MedAE (s) & MedRE \\
\hline
\hline
\lr & 194.604 & 0.2949 & 164.820 & 0.3017 \\
\hline
\hline
\avg & 178.459 & 0.2704 & 120.834 & 0.2345 \\
\hline
\hline
\temprel & 149.815 & 0.2270 & 97.365 & 0.1907 \\
\hline
\temparima & 143.311 & 0.2171 & 98.780 & 0.1890 \\
\hline
\hline

\temprelR & 143.719 & 0.2178 & \textbf{92.067} & \textbf{0.1805} \\
\hline
\temparimaR & \textbf{142.334} & \textbf{0.2157} & 98.046 & 0.1874 \\
\hline
\hline
\multicolumn{5}{|l|}{Comparison with Bing Maps (260K testing trips)} \\
\hline
\bing & 202.684 & 0.3157  & 134.000 & 0.2718 \\
\hline
\bingtraffic & 242.402 & 0.3776 & 182.000 & 0.3395 \\
\hline
\temparimaR & \textbf{135.365} & \textbf{0.2108} & \textbf{94.940} & \textbf{0.1839} \\
\hline
\end{tabular}
\end{center}
\end{table}%

We first observe that our method is better than the linear regression (\lr) baseline. This is expected because \lr is a simple baseline which does not consider the origin and destination locations. 

Considering temporal variations of traffic condition improves the estimation performance. All the \temp methods have significantly lower error compared with the \avg method. The improvement of \temparima over \avg is about $35$ seconds in MAE. In other words, the MAE is decreased by nearly $20\%$ by considering the temporal factor.  Furthermore, using the absolute speed reference (\temparima) is better than using the relative (\ie, weekly) speed reference (\temprel). This is because the traffic condition does not strictly follow the weekly pattern, but has some irregular days such as holidays.

By adding the region factor, the performance is further improved. This means the traffic pattern between regions are actually different.  However, we observe that the improvement of  \temparimaR over \temparima is not as significant as the improvement of \temprelR over \temprel. This is due to data sparsity when we compute the absolute time reference for each region pair. If two regions have little traffic flow, the absolute time reference might not be accurate. In this case, using the relative time reference between regions make the data more dense and produces better results.

Comparing with \bing on the $260k$ testing trips, \temparimaR significantly outperforms \bing by $67$ seconds. \bing underestimates the trips without considering traffic, where $64.53\%$ testing trips are underestimated. However, when considering traffic (the query was sent to the API at the same time and the same day of the week), $75.02\%$  testing trips are overestimated.

\begin{figure}[t]
\centering
\subfigure[MAE w.r.t. training set size\label{fig:sample-mae}]{\includegraphics[width=0.23\textwidth]{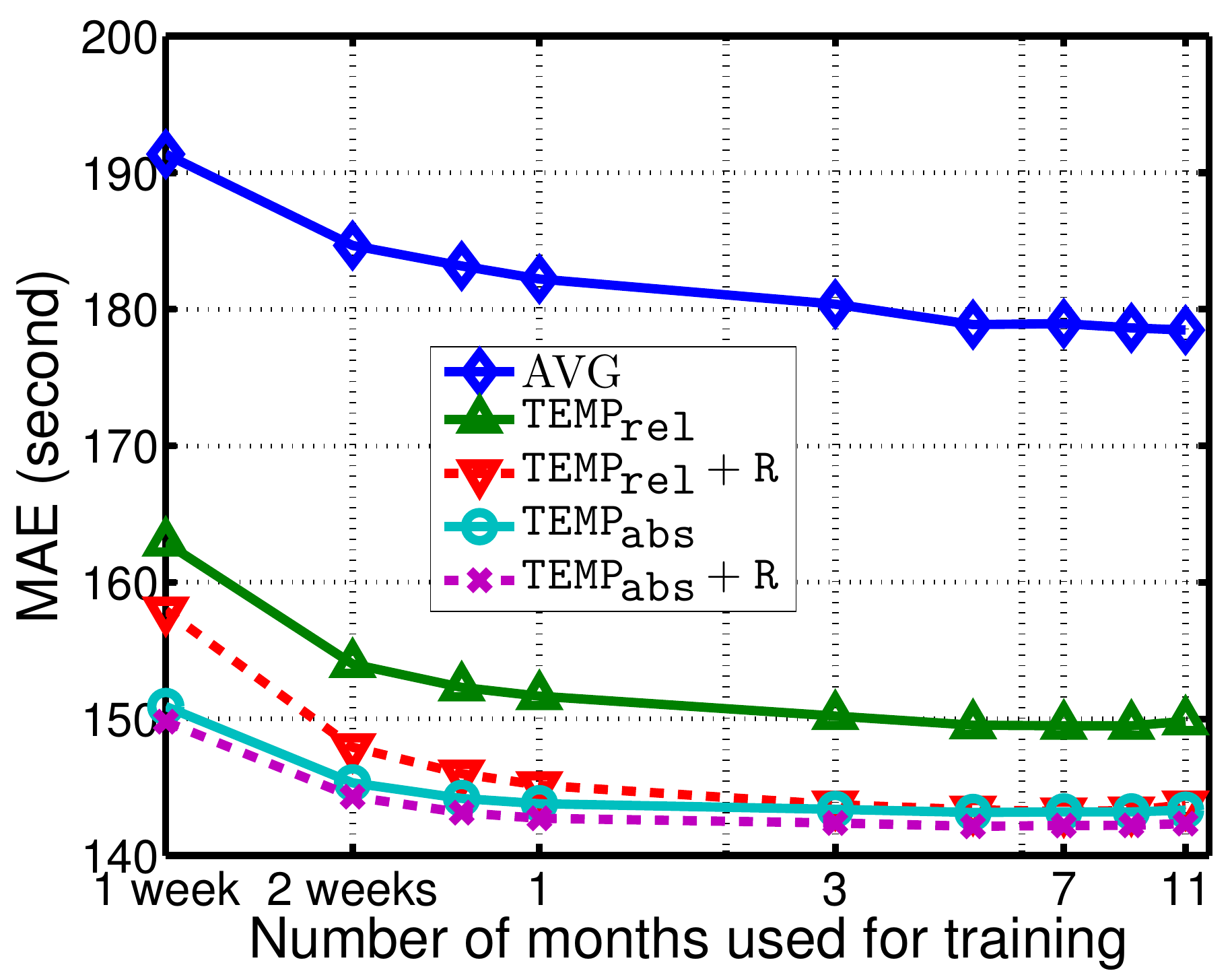}}
\subfigure[Coverage of testing set\label{fig:sample-cover}]{\includegraphics[width=0.23\textwidth]{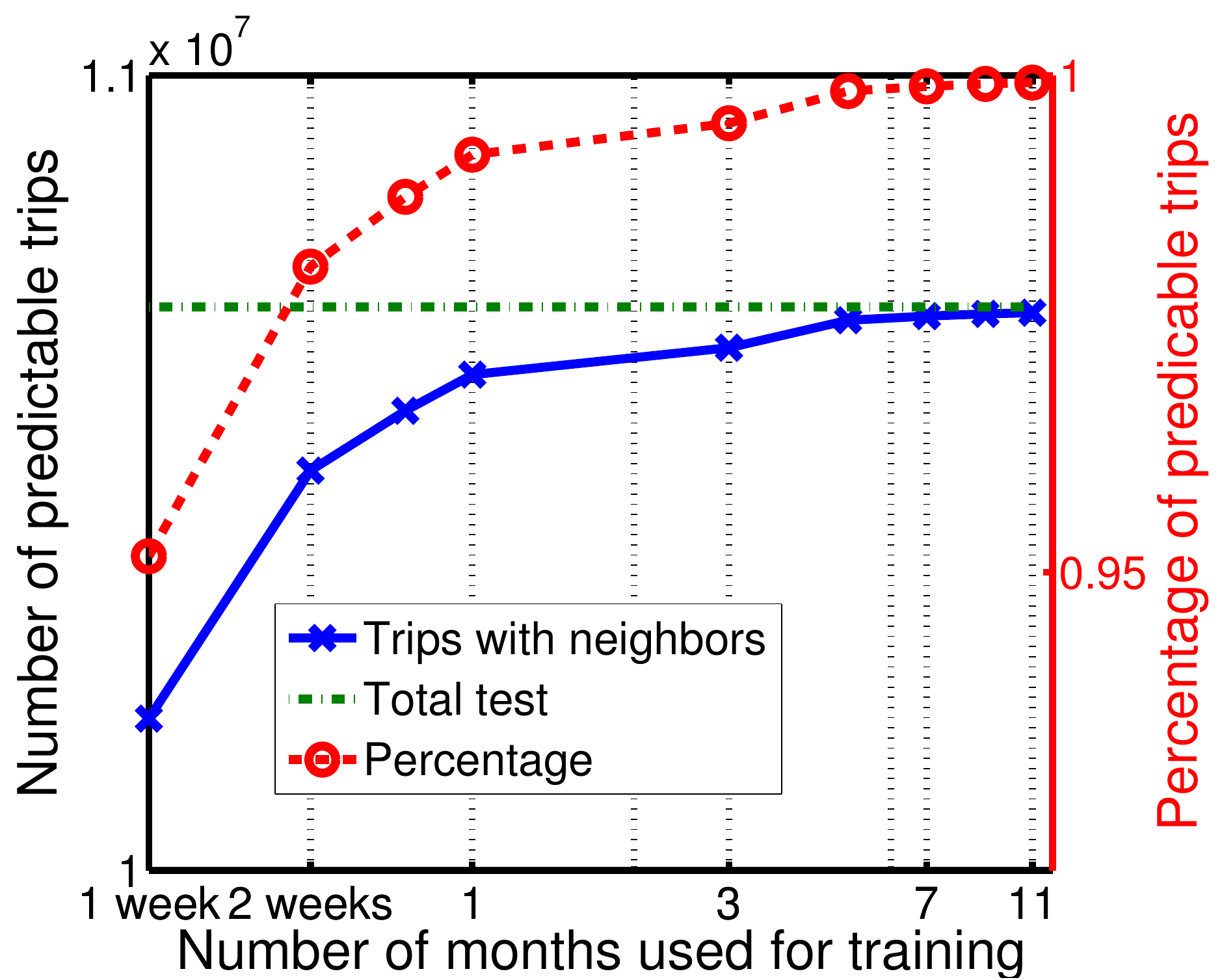}}
\caption{Performance with regard to the training set size.}\label{fig:sample-training}
\end{figure}

\subsubsection{Performance w.r.t. the Size of Historical Data}

We expect that, by using more historical data (\ie, training data), the estimation will be more accurate. To verify this, we study how performance changes w.r.t. the size of training data. We choose different time durations (from 1 week, 2 weeks, to 11 months) before December as the training data. Figure~\ref{fig:sample-mae} shows the accuracy w.r.t. the size of the training data. The result meets with our expectation that MAE drops when using more historical data. But the gain of using more data is not obvious when using more than 1 month data. This indicates that using 1-month training data is enough to achieve a stable performance in our experimental setting. Such indication will save running time in real applications as we do not need to search for neighboring trips more than 1 month ago.

In addition, using more historical data, we are able to cover more testing trips. Because a testing trip should have at least one neighboring trip in order to be estimated. Figure~\ref{fig:sample-cover} shows the coverage of testing data by using more historical data. With 1-week data, we can estimate $95.16\%$ testing trips; with 1-month data,  we can cover $99.20\%$ testing trips.

\nop{One of our fundamental assumption is that the more historical data we have, the better we can estimate the travel time, and we should be able to estimate more trips as well.

To evaluate this assumption, we select different subsets of the training set, and estimate the December trips respectively. More specifically, we select duration which is  right before the December, and then extract all the trips in this duration as training. The durations we picked are 1 week, 2 weeks, 3 weeks, 1 month, $\cdots$, 11 months. Evaluation results are shown in Figure~\ref{fig:sample-training}.

In Figure~\ref{fig:sample-mae}, we show the MAE of various methods against the number of training trips we selected. Overall, the MAE will decrease as the training set size increases. We notice that even with only 1 week of training trips, our \temparimaR can achieve $149.79$ seconds error. Beyond one month, more training trips improve the estimation marginally, only from $142.72$ to $142.20$. 

In Figure~\ref{fig:sample-cover}, the number of predictable testing trips is plotted against the training set size. We show the actual number of testing trips with solid blue line, additionally we plot the total number of testing trips in December as reference (green line). We also plot the percentage of predictable trips on the right y-axis. Even with only 1-week training trips, we can estimate $95.16\%$ testing trips. With more training trips, we are able to estimate more testing trips.
}

\subsubsection{Performance w.r.t. trip features.}

\begin{figure}[t]
  \centering
  \subfigure[MAE w.r.t. travel time\label{fig:t-mae}]{\includegraphics[width=0.23\textwidth]{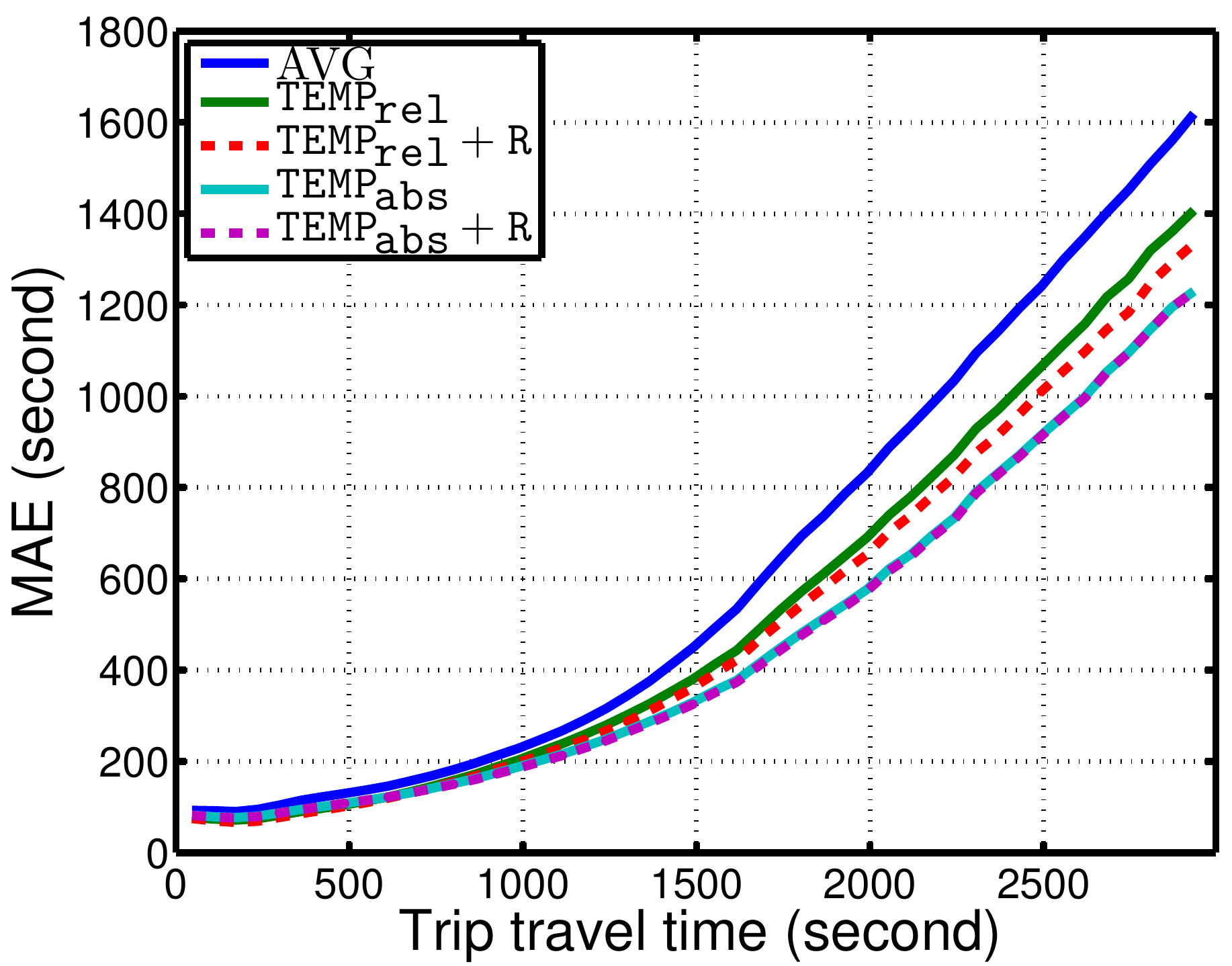}}
  \subfigure[MRE w.r.t. travel time\label{fig:t-mre}]{\includegraphics[width=0.23\textwidth]{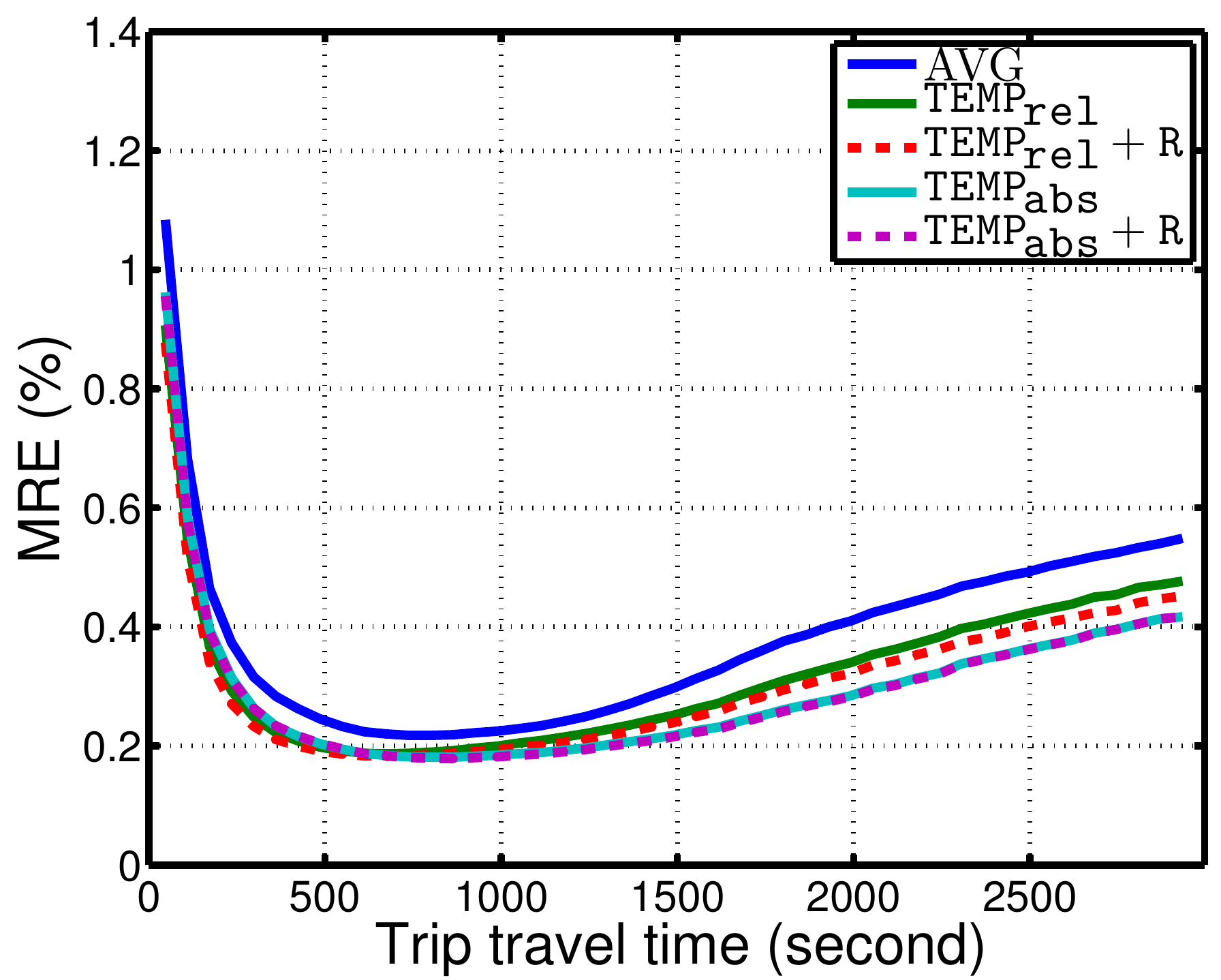}}

  \subfigure[MAE w.r.t. distance\label{fig:d-mae}]{\includegraphics[width=0.23\textwidth]{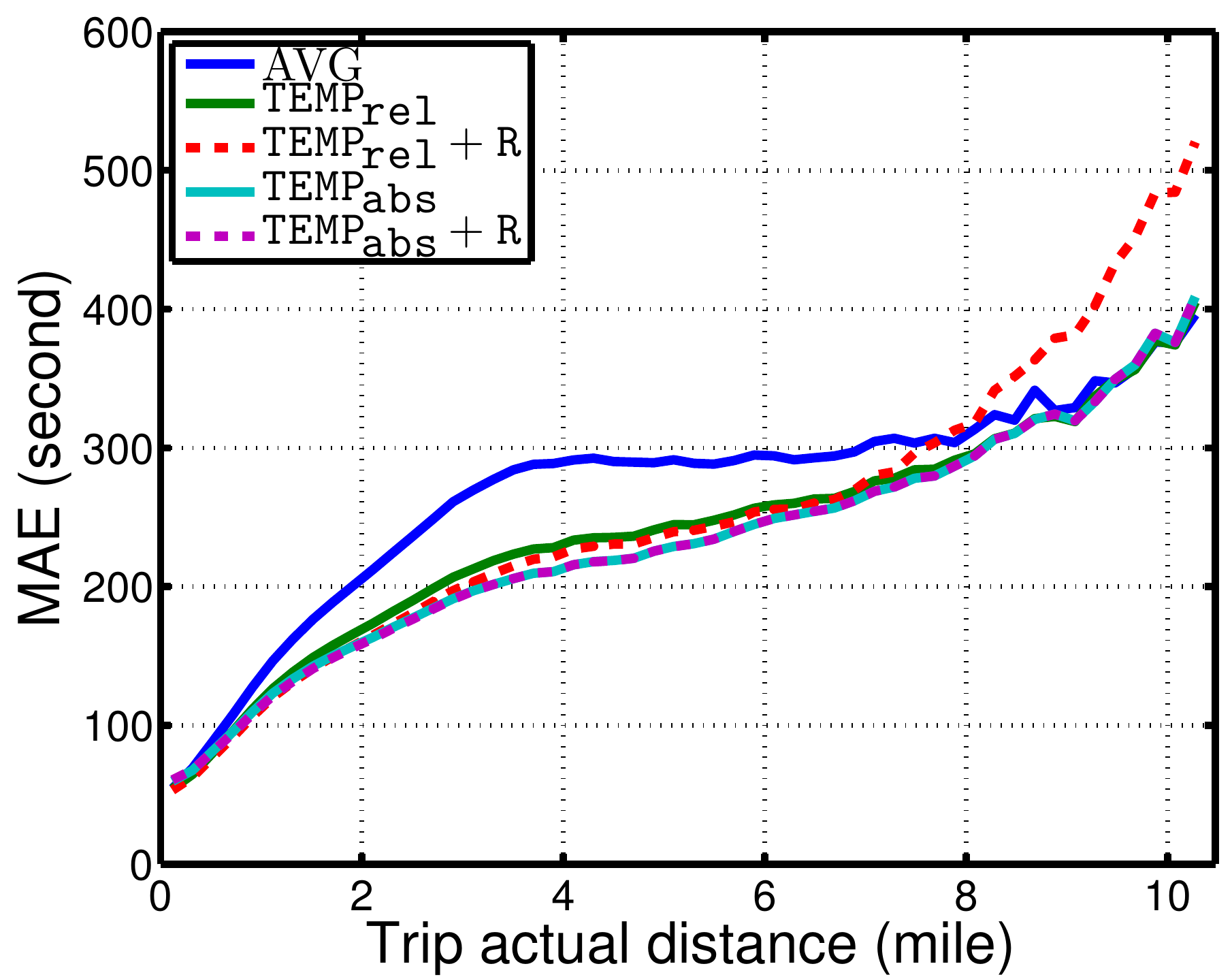}}
  \subfigure[MRE w.r.t. distance\label{fig:d-mre}]{\includegraphics[width=0.23\textwidth]{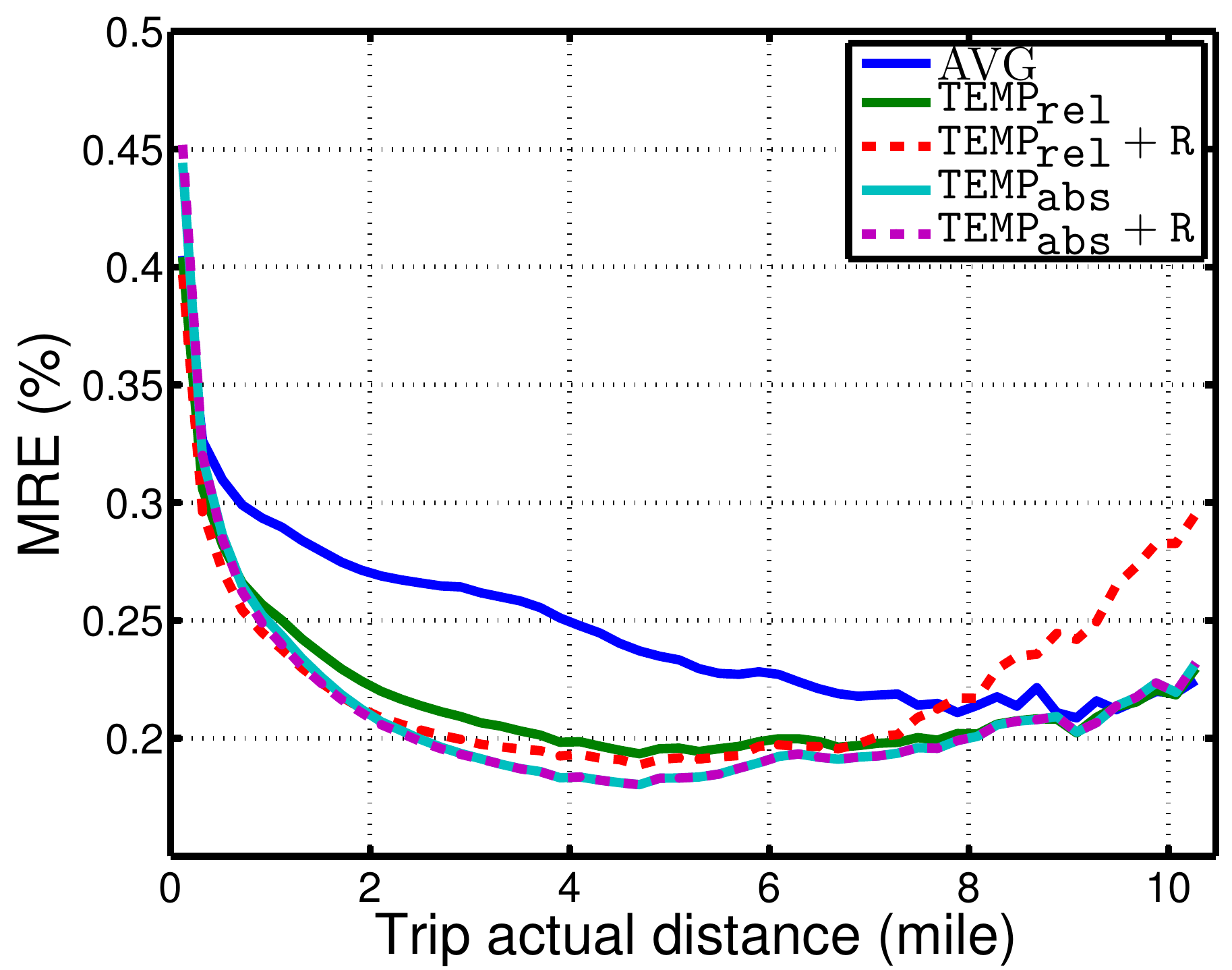}}

  \subfigure[MAE w.r.t. number of neighbors\label{fig:nn-mae}]{\includegraphics[width=0.23\textwidth]{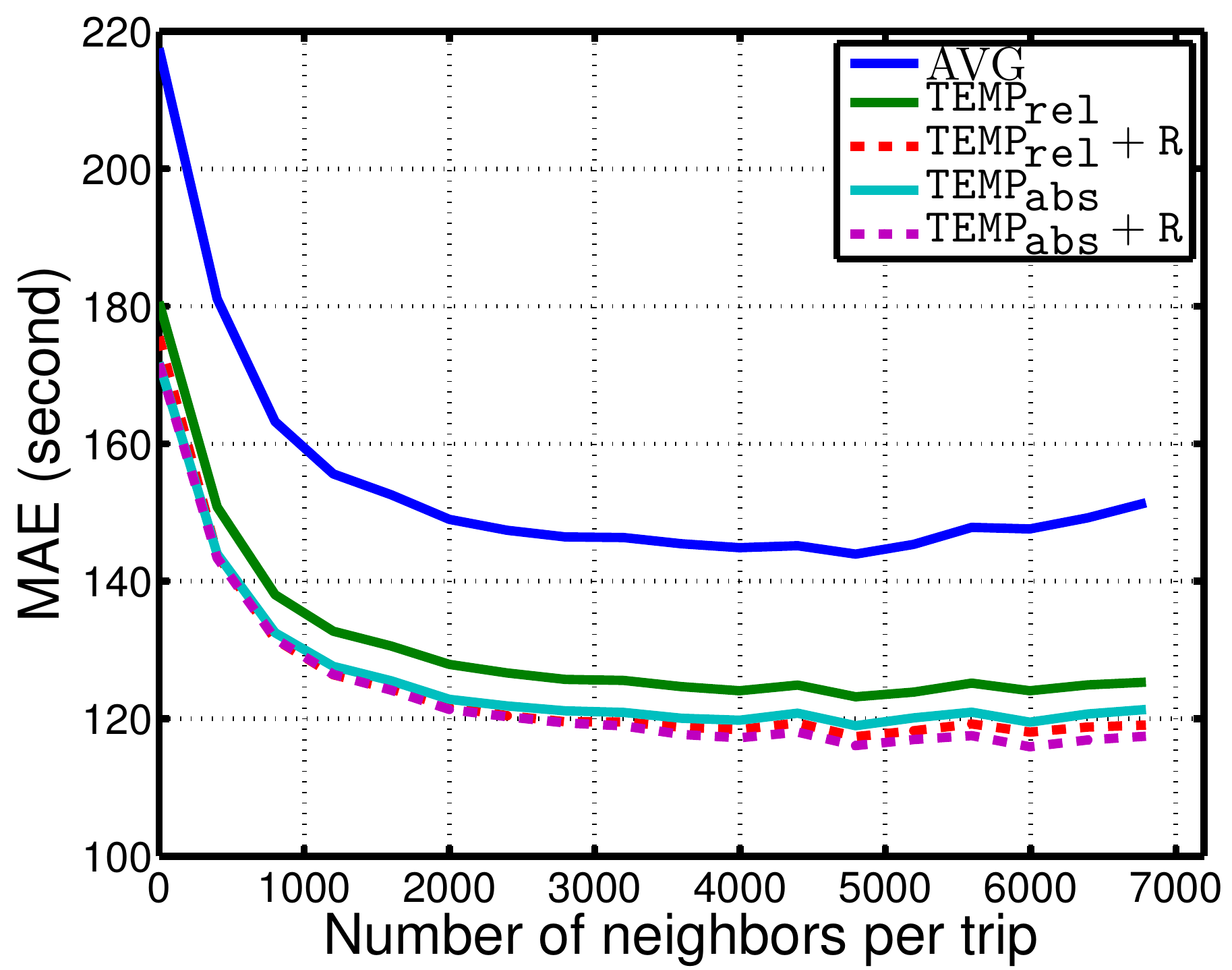}}
  \subfigure[MRE w.r.t. number of
  neighbors\label{fig:nn-mre}]{\includegraphics[width=0.23\textwidth]{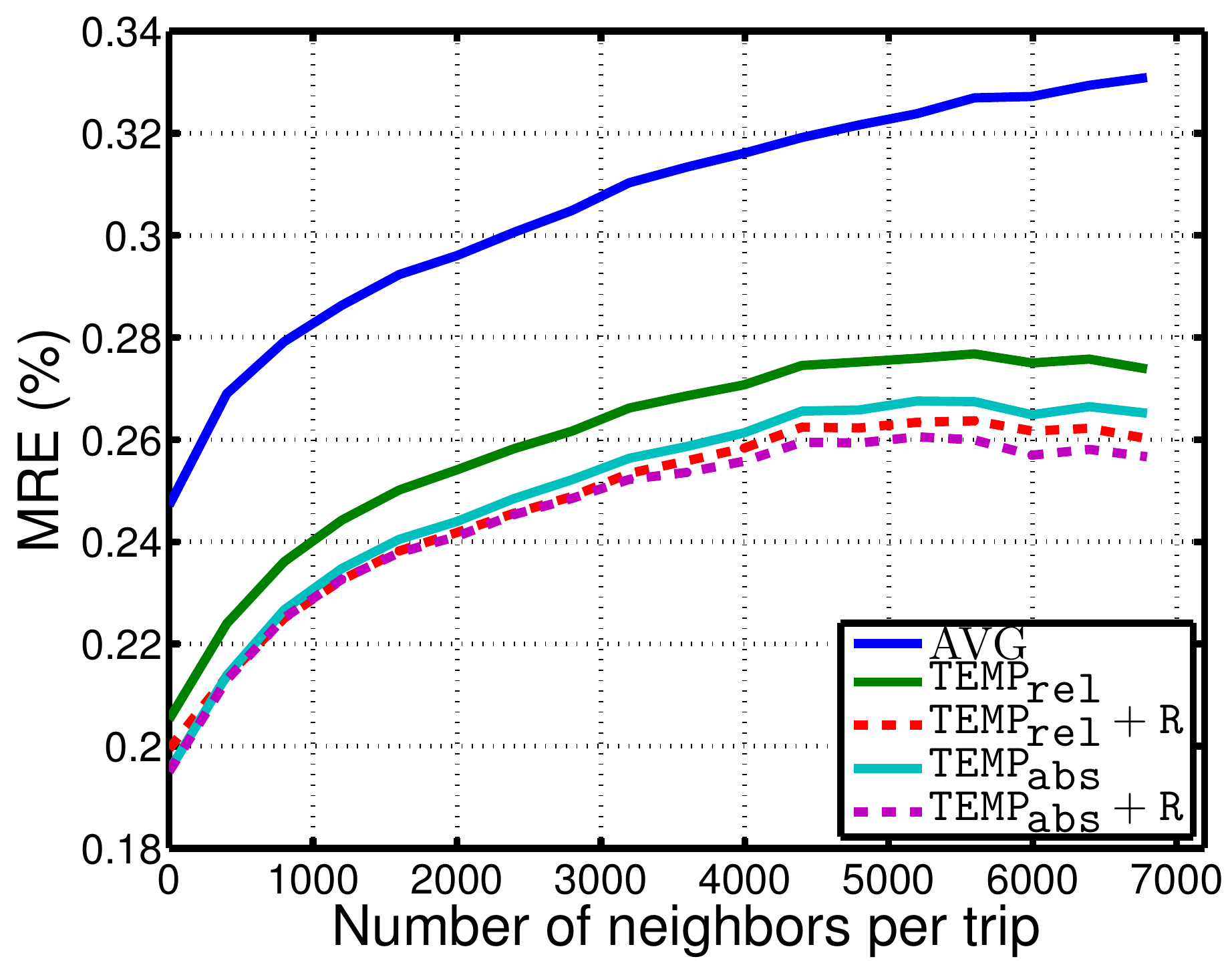}}
\caption{Performance with regard to the trip features.}\label{fig:mae-data}
\end{figure}

 \textbf{Performance w.r.t. trip time.} In Figure~\ref{fig:t-mae} and \ref{fig:t-mre}, we plot the MAE and MRE with respect to the trip time. As expected, the longer-time trips have higher MAEs. It is interesting to observe that the MREs are high for both short-time trips and long-time trips. Because the short-time trips (less than $500$ seconds) are more sensitive to dynamic conditions of the trips, such as traffic light. On the other hand, the long-time trips usually have less neighbors, which leads to higher MREs.

\textbf{Performance w.r.t. trip distance.}  The MAE and MRE against trip distance are shown in Figure~\ref{fig:d-mae} and \ref{fig:d-mre}. Overall, we have consistent observations as Figure~\ref{fig:t-mae} and Figure~\ref{fig:t-mre}.  However, we notice that the error of \temprelR increases faster than other methods. \temprelR becomes even worse than \avg method when the trip distance is longer than 8 miles. This is because the longer-distance trips have fewer neighboring trips. However, we do not observe the same phenomenon in Figure~\ref{fig:t-mae} and Figure~\ref{fig:t-mre}. After further inspection, we find that long-\emph{distance} trips have even less neighboring trips than long-\emph{time} trips. The reason is that long-distance trips usually have long travel time; however, long-time trips may not have long trip distance (e.g., the long travel time may be due to traffic). 

\nop{Two remarks should be mentioned here. 1) The trips with long travel times does not necessarily have long actual distances, which is why we did not observe this phenomenon in the MAE against travel time plot. 2) The \temparimaR faces this same problem, together with another sparseness issue, and we use \temparima to fill in the missing values in \temparimaR to address the sparseness issue. Therefore, for longer trips the \temparimaR degrades and is asymptotic to \temparima.}

\textbf{Performance w.r.t. number of neighboring trips.} The error against number of neighbors per trip is shown in Figure~\ref{fig:nn-mae} and Figure~\ref{fig:nn-mre}. MAEs decrease for the trips with more neighbors. However, MREs increase as the number of neighbors increases. The reason is that the short trips usually have more neighbors and short trips have higher relative errors, as shown in Figure~\ref{fig:t-mre} and Figure~\ref{fig:d-mre}.

\subsubsection{An Alternative Definition of Neighbors}
\label{subsec:nyc-spatial-weight}

As discussed in Section~\ref{sec:method:problem:over}, now we study the performance by defining \emph{soft} neighbors.  We give neighboring trips spatial weights based on the sum of corresponding end-points' distance, and evaluate the performance on $500k$ trips. The MAE of \temparimaR is $107.705$s on this testing set. Adding spatial weights on the neighbors degenerates the performance to $112.392$s. It suggests that it is non trivial to assign weights on neighbors based on the distances of end-points. It will require more careful consideration of this factor.


\input{4experiment-sh.tex}

\subsection{The Impacts of Outliers}
\label{sec:exp-outliers}

We implemented several outlier filters to work as a pipeline. For NYC data, we have filters using time and distance, time and fare, distance and fare, trip distance and distance between end points, trip time and distance between end points, respectively. In total, an estimated $7\%$ outliers are found.
\nop{
Real world data often contain a large number of outliers. We observe many outlying trips in our taxi trip datasets (e.g., a trip with distance between starting and ending point as only one mile has an hour long travel time). If we do not filter these outliers properly,}

The outliers will impact the experimental performances in two ways: (1) the outliers in the training data will make the travel time estimation less accurate; and (2) the outliers in the testing trips will make the experimental evaluations less reliable because the errors of the outlying testing trips will greatly affect the overall accuracy. 

\begin{table}[h]
\caption{Performance with/without Outlier filtering}\label{tab:filt-out}
\centering
\begin{tabular}{|c||c|c||c|c||c|c|}
\hline
 & \multicolumn{2}{|c||}{Setting 1} & \multicolumn{2}{|c||}{Setting 2} & \multicolumn{2}{|c|}{Setting 3}  \\ \hline
Training set & \multicolumn{2}{|c||}{without outliers} & \multicolumn{2}{|c||}{with outliers} & \multicolumn{2}{|c|}{with outliers}  \\ \hline
Testing set & \multicolumn{2}{|c||}{without outliers} & \multicolumn{2}{|c||}{without outliers} & \multicolumn{2}{|c|}{with outliers}  \\ \hline\hline
Method & MAE & MRE & MAE & MRE & MAE & MRE \\ \hline \hline
\lr         & 212.704 & 0.3238& 217.18  & 0.3305  & 232.605 & 0.3484 \\ \hline
\avg        & 182.215 & 0.2774& 186.733 & 0.2841  & 207.439 & 0.3107 \\ \hline
\temprel    & 151.667 & 0.2309& 155.711 & 0.2369  & 177.587 & 0.2660 \\ \hline
\temprelR   & 145.149 & 0.2210& 149.101 & 0.2269  & 171.091 & 0.2563 \\ \hline
\temparima  & 143.794 & 0.2189& 148.039 & 0.2253  & 170.036 & 0.2547 \\ \hline
\temparimaR & 142.731 & 0.2173& 148.039 & 0.2253  & 170.036 & 0.2547 \\ \hline
\end{tabular}
\end{table}

We study how outliers impact these two aspects using NYC dataset with November trips as training and December trips as testing. The number of anomalous trips in November and December are $758,253$ out of $11,915,496$ (i.e., $6.36\%$) and $801,503$ out of $11,434,273$ (i.e., $7.00\%$), respectively. Three different experiment settings are selected as shown in Table~\ref{tab:filt-out}. Comparing the results of the first setting with the second setting, we can see that by taking out $6.4\%$ outliers from training trips, the MAE is reduced by roughly $5$ seconds. It is critical to note that the outliers in testing set will incur even a very large error in the evaluation.  In setting 3 of Table~\ref{tab:filt-out}, where both testing and training set have outliers,  MAE is more than $20$ seconds worse than the first setting.

\subsection{Running Time}
\label{sec:running-time}

Table~\ref{fig:runtime} presents the running time comparison between \temparimaR and \subpath. We profile the running time of \temparimaR on last two months' trips in NYC dataset, and \subpath on whole Shanghai dataset. Note that NYC dataset is bigger than Shanghai dataset and \temparimaR is the most computationally expensive method, and also the most accurate method among our proposed methods. The pre-processing time for \temparimaR consists of mapping trips to grids, calculating the speed reference, and learning the ARIMA parameters. The preparation time for \subpath  includes indexing trajectory by road segments.

\begin{table}[h!]
\centering
\caption{Running time comparison}\label{fig:runtime}
\begin{tabular}{|c|c|c|c|c|}
\hline
Method & Process & \#trips & time & time/trip \\ \hline
{\multirow{6}{*}{\temparimaR}} & {\pbox[b]{2.7cm}{Prepare training trip}} & $11$ million & $42$s & $0.0038$ms \\ \cline{2-5}
& \pbox[b]{2.7cm}{Learn ARIMA} & $11$ million & $147$s & $0.0133$ms \\ \cline{2-5}
& \pbox[b]{2.7cm}{Find neighbor (1 thread)} & $100$k & $109$s & $1.09$ms \\ \cline{2-5}
& \pbox[b]{2.7cm}{Find neighbor (8 threads)} & $100$k & $24$s & $0.24$ms \\ \cline{2-5}
& \pbox[b]{2.7cm}{Estimate (1 thread)} & $10.5$ million & $6560$s & $0.065$ms \\ \cline{2-5}
& \pbox[b]{2.7cm}{Estimate (8 threads)} & $10.5$ million & $229$s & $0.0208$ms \\ \hline
{\multirow{2}{*}{\subpath}} & \pbox[b]{2.7cm}{Prepare training trip} & $4.35$ million & $547$s & $0.126$ms \\ \cline{2-5}
& \pbox[b]{2.7cm}{Estimate (1 thread)} & $54,503$ & $2297$s & $42$ms \\\hline
\end{tabular}
\end{table}

In \temparimaR, both training trip preparation and learning ARIMA can be done offline or updated every hour with the new data. The online query part includes querying neighboring trips and estimation using neighbors. The neighbor finding is the most time consuming part. We get $1.09$ ms/trip if using one thread and $0.24$ ms/trip if using 8 threads. The estimation part only takes $0.02$ ms with $8$ threads. So the total online query time for our method is  $1.505$ ms/trip on average and it will be even faster (i.e., $0.26$ ms/trip) if using 8 threads. The estimation of \subpath includes finding optimal concatenation of segments, which has $O(n^2\cdot m)$ time complexity, where $n$ is the number of road segments and $m$ is number of trips going through each segment. It takes 42ms on average to estimate a testing trip. Our neighbor-based method is more efficient in answering travel time queries because we avoid computing the route and estimating the time for the subpaths.


%% file: 4experiment-sh.tex

\subsection{Performance on Shanghai Data}
\label{sec:exp:shanghai}
In this section, we conduct experiments on Shanghai taxi data. This dataset provides the trajectories of the taxi trips so we can compare our method with \seg and \subpath methods.
Both \seg and \subpath methods use the travel time on individual road segments or subpaths to estimate the travel time for a query trip. Due to data sparsity, we cannot obtain travel time for every road segments. Wang et al.~\cite{WZX14} propose a tensor decomposition method to estimate the missing values. In our experiment, to avoid the missing value issue, we only select the testing trips that have values on every segments of the trip. However, by doing such selection, the testing trips are biased towards shorter trips. Because the shorter the trip is, the less likely the trip has missing values on the segment. To alleviate this bias issue, we further sampled the biased trip dataset to make the travel time distribution similar to the distribution of the original whole dataset. The testing set contains $2,138$ trips in total.

\begin{table}[ht!]
\centering
\caption{Overall Performance on Shanghai Data}\label{tab:comp-sh}
\begin{tabular}{|l|l|l|l|l|}
\hline
Method & MAE & MRE & MedAE & MedRE \\
\hline\hline
\lr & 130.710 & 0.6399 & 138.796 & 0.6173 \\
\hline
\baidu & 111.484 & 0.5451 & 73.001 & 0.5001 \\
\hline\hline
\seg & 119.833 & 0.5866 & 84.704 & 0.4947 \\
\hline
\subpath & 113.566 & 0.5560 & 75.913 & 0.4820 \\
\hline\hline
\avg & 94.202 & 0.4615 & 60.183 & 0.3739 \\
\hline
\temprel & \textbf{92.428} & \textbf{0.4527} & \textbf{55.317} & \textbf{0.3678} \\
\hline
\end{tabular}
\end{table}

\subsubsection{Overall Performance on Shanghai Data}

The comparison among different methods is shown in Table~\ref{tab:comp-sh}. Our neighbor-based method significantly outperform other methods. The simple method \avg is 17 seconds better than \baidu in terms of MAE.  By considering temporal factors, method \temprel and \temparima further outperform \avg method. \subpath method outperforms \seg method, which is consistent with previous work~\cite{WZX14}.  \seg simply adds travel time of individual segments, whereas \subpath better considers the transition time between segments by concatenating subpaths.  However, \subpath is still 21 seconds worse than our method \temprel method in terms of MAE. Such result is also expected. Because our method can be considered as a special case \subpath method, where we use the whole paths from the training data to estimate the travel time for a testing trip. If we have enough number of whole paths, it is better to use the whole paths instead of subpaths. 

Since the Shanghai taxi data are much more sparse than NYC data, we do not show the results of \temprelR and \temparimaR on Shanghai dataset. Additionally, the two-month Shanghai data have several days' trip missing, and the ARIMA estimation require a complete time series. Therefore, we did not show the results of \temparima, either. The main idea here is to show that \temprel can outperform the existing methods, and with more data our other approaches should outperform the \temprel as well as shown in NYC data.

\subsubsection{Applicability of Segment-based Method and Neighbor-based Method}
Segment-based method requires every individual road segment of the testing trip has at least one historical data point to estimate the travel time. On the other hand, our neighbor-based method requires a testing trip has at least one neighboring trip. Among $435,887$ trips in Shanghai dataset, only $54,530$ trips have values on every road segment, but $217,585$ trips have at least one neighbor. Therefore, in this experimental setting, our method not only outperforms segment-based method in terms of accuracy, it is also more applicable to answer more queries (\textbf{$49.9\%$}) compared with segment-based method ($12.5\%$).

%% file: 6conclusion.tex
\section{Conclusion}
\label{sec:conclusion}

This paper demonstrates that one can use large-scale trip data to estimate the travel time between an origin and a destination in a very efficient and effective way. Our proposed method retrieves all the neighboring historical trips with the similar origin and estimation locations and estimate the travel time using those neighboring trips. We further improve our method by considering the traffic conditions w.r.t. different temporal granularities and spatial dynamics. In addition, we develop an outlier filtering method to make our estimation more accurate and experimental results more reliable. Finally, we conduct experiments on two large-scale real-world datasets and show that our method can greatly outperform the state-of-the-art methods and online map services. 
In the future, we are interested in finding out what  other new insights could be offered by such big data, for example, how crime and economic status of places correlate with the travel flows.